\documentclass{article}

\usepackage{arxiv}

\usepackage[utf8]{inputenc}
\usepackage[T1]{fontenc}
\usepackage{hyperref}
\usepackage{url}
\usepackage{booktabs}
\usepackage{amsmath,amssymb}
\usepackage{algorithm}
\usepackage{algpseudocode}
\usepackage{amsfonts}
\usepackage{nicefrac}
\usepackage{microtype}
\usepackage{graphicx}
\usepackage{natbib}
\usepackage{doi}
\usepackage{longtable}
\usepackage{array}
\usepackage{tikz}
\usetikzlibrary{shapes.geometric, arrows.meta, positioning}

\title{Counterfactual Bias Testing for Application Tracking Systems}

\author{
Sai Yashwant \\
AI Product \& Platform Head \\
ManpowerGroup Services India Pvt. Ltd. \\
\texttt{sai.yashwant@manpowergroup.com}
\And
Shruti Bansal \\
Senior Data Scientist \\
ManpowerGroup Services India Pvt. Ltd. \\
\texttt{shruti.bansal@manpowergroup.com}
\And
Anurag Dubey \\
Data Scientist \\
ManpowerGroup Services India Pvt. Ltd. \\
\texttt{anurag.dubey@manpowergroup.com}
\And
Samaroha Chatterjee \\
Data Scientist \\
ManpowerGroup Services India Pvt. Ltd. \\
\texttt{samaroha.chatterjee@manpowergroup.com}
\And
Satyam Kumar \\
Data Scientist \\
ManpowerGroup Services India Pvt. Ltd. \\
\texttt{satyam.kumar@manpowergroup.com}
\And
Shreyash Gupta \\
Data Scientist \\
ManpowerGroup Services India Pvt. Ltd. \\
\texttt{shreyash.gupta@manpowergroup.com}
\And
Gantala Thulsiram \\
Assistant Professor \\
Indian Institute of Technology, Hyderabad \\
\texttt{thulsiramg@mae.iith.ac.in}
}

\renewcommand{\shorttitle}{Counterfactual Bias Testing for Application Tracking Systems}
\renewcommand{\headeright}{}
\renewcommand{\undertitle}{}

\hypersetup{
pdftitle={Counterfactual Bias Testing for Application Tracking Systems},
pdfsubject={cs.CY, cs.CL, cs.IR},
pdfauthor={Sai Yashwant, Shruti Bansal, Anurag Dubey, Samaroha Chatterjee, Satyam Kumar, Shreyash Gupta, Gantala Thulsiram},
pdfkeywords={Algorithmic Hiring Bias, Correspondence Audit, Counterfactual Fairness, Synthetic Data Generation, LLM Agents, Fairness Metrics, EU AI Act},
}

\begin{document}
\maketitle

\begin{abstract}
Automated candidate--job matching systems are increasingly classified as high-risk AI under emerging regulation, yet auditing them for demographic bias is expensive: classical correspondence-audit studies require hand-crafted resume pairs and manual submission, which does not scale to the rate at which enterprise matching pipelines are retrained and redeployed. This paper presents a general, reusable methodology that (1)~uses a chain of task-specialized large-language-model (LLM) agents to synthesize identity-neutral base resumes and inject controlled demographic treatments across five protected-characteristic axes (sex/gender, age, place of residence, language, disability signal), producing a correspondence-audit-style $K\times(1+N)$ matrix of base candidates and bias variants; (2)~qualitatively flags inferred protected characteristics and bias per an EU AI Act-aligned prompt; (3)~ranks every candidate against a job description using a fine-tuned sentence-embedding model and cosine similarity, representative of the semantic ranking core used by modern candidate--job matching systems; and (4)~computes a nine-metric quantitative fairness suite -- spanning counterfactual (score delta, mean absolute rank change, flip rate), group-fairness (top-$K$ retention, four-fifths rule or impact ratio), and merit-aware (Recall@$K$, nDCG@$K$, equal opportunity, equalized odds) families -- each with bootstrap confidence intervals, exact/asymptotic significance tests (Wilcoxon signed-rank, McNemar, Fisher exact, Wilson score), and Benjamini--Hochberg false-discovery-rate correction, culminating in an automatically generated PASS/INVESTIGATE/FAIL audit report with a composite risk score. We illustrate the methodology on an example correspondence-audit corpus spanning $5$ job orders, $100$ base candidates, and $10$ demographic-bias treatments ($90$ metric$\times$variant evaluations), illustrating that while variant-level score shifts, top-$K$ retention, and merit-aware true-positive-rate/false-positive-rate gaps remain within tolerance for every treatment, a rank-stability metric (mean absolute rank change) and a ranking-quality metric (nDCG@$K$) each surface borderline findings -- including one on the neutral baseline configuration itself -- that a score- or retention-only view would have missed. The results argue for multi-metric, multi-family auditing over any single aggregate fairness score, and for treating LLM-agent-generated correspondence audits as a practical, low-cost complement to human-curated audit studies for any candidate--job matching pipeline.
\end{abstract}


\section{Introduction}
\label{sec:intro}

Automated candidate--job matching, encompassing resume parsing, semantic ranking, and shortlist generation, has moved from an experimental augmentation of recruiter workflow to a load-bearing component of high-volume staffing pipelines, with contextual-transformer encoders~\citep{devlin2019bert} and their fine-tuned derivatives~\citep{zhang2020bert,qin2018enhancing} now forming the semantic core of many such systems. Under the European Union's Artificial Intelligence Act, AI systems used ``to evaluate candidates... in the recruitment or selection of natural persons'' are explicitly enumerated as high-risk~\citep{eu2024aiact}, and under longstanding United States employment law, a selection procedure that produces a substantially different selection rate across protected groups is presumptively suspect under the four-fifths rule codified in the Uniform Guidelines on Employee Selection Procedures~\citep{eeoc1978uniform}. These are not abstract concerns: field experiments going back to \citet{bertrand2004emily} demonstrated that resumes with distinctively White-sounding names received substantially more callbacks than otherwise-identical resumes with distinctively Black-sounding names, and more recent audits of commercial hiring and screening tools have repeatedly found measurable disparate treatment along gender, race, age, and disability lines~\citep{dastin2018amazon,bogen2018help,raghavan2020mitigating,ajunwa2020paradox,sanchezmonedero2020solve,wilson2024gender}. As matching systems increasingly delegate their semantic core to large language models (LLMs) and LLM-derived embeddings, the risk surface changes shape but does not shrink: LLMs inherit and can amplify societal biases present in their training corpora~\citep{gallegos2024bias,bommasani2021opportunities,nadeem2021stereoset,buolamwini2018gender}, and a matching pipeline built on such components requires the same scrutiny as any other automated hiring tool.

The traditional instrument for measuring hiring discrimination is the \emph{correspondence audit}: construct pairs (or larger matrices) of resumes that are equivalent in every job-relevant respect but differ in one or more protected-characteristic signals, submit them through the real or simulated selection process, and measure differential treatment~\citep{bertrand2004emily,kim2017datadriven}. Correspondence audits are the gold standard for causal claims about discrimination because the counterfactual comparison is built into the design, but they are also expensive: constructing dozens of matched resume variants per job family, and repeating the exercise every time a matching model is retrained, does not fit the release cadence of a modern MLOps pipeline. At the same time, the machine-learning fairness literature has produced a rich vocabulary of quantitative metrics -- demographic/statistical parity, equalized odds and equal opportunity~\citep{hardt2016equality}, disparate impact and its remover~\citep{feldman2015certifying}, counterfactual fairness~\citep{kusner2017counterfactual}, and ranking-specific notions of exposure and top-$K$ fairness~\citep{singh2018fairness,yang2017measuring,zehlike2017fair} -- together with general-purpose toolkits such as AI Fairness 360~\citep{bellamy2018aif360}, Fairlearn~\citep{bird2020fairlearn}, and Aequitas~\citep{saleiro2018aequitas}. These toolkits, however, assume a labeled classification or regression task and a real deployed population; they do not natively support the paired, counterfactual, per-treatment structure that a correspondence audit produces, nor do they generate the audit data itself.

The methodology presented in this paper is designed to close this gap by treating correspondence-audit \emph{generation} and \emph{quantitative fairness evaluation} as two halves of the same pipeline. On the generation side, a chain of task-specialized LLM agents (i)~elicits a list of protected-characteristic bias descriptors, (ii)~synthesizes a set of identity-neutral base candidate resumes together with a matrix of bias-variant resumes that inject exactly one demographic treatment axis at a time while holding qualifications fixed, and (iii)~produces an EU AI Act-aligned qualitative bias flag for every resume. On the evaluation side, every accumulated candidate identity is ranked against a job description using a structuring-plus-embedding-plus-cosine-similarity pipeline representative of the semantic ranking core used in modern candidate--job matching systems, and the resulting paired baseline/variant rankings are passed through a nine-metric fairness suite spanning counterfactual, group-fairness, and merit-aware families, each accompanied by a confidence interval, a significance test appropriate to its estimator, and a Benjamini--Hochberg multiple-testing correction~\citep{benjamini1995controlling}. The output is a single, self-contained HTML report, styled as an executive audit dashboard, with a composite PASS/INVESTIGATE/FAIL classification per metric-and-variant and an aggregate weighted risk score.

We organize the methodology around three research questions that any correspondence-audit-style bias test of a matching system must answer. \textbf{RQ1:} Does injecting only identity or protected-characteristic text into a candidate profile, holding all qualifications fixed, shift the candidate's relevance score or rank against a job? \textbf{RQ2:} Does such injection shift candidates across an operational shortlisting cutoff (top-$K$), changing who is actually advanced? \textbf{RQ3:} Where shifts occur, does their direction disproportionately disadvantage specific protected groups, as opposed to merely affecting them? The distinction embedded in RQ3 is maintained throughout this paper: bias is defined relative to the group that is deprioritized or disadvantaged, not the group that is favored (Section~\ref{sec:method-group}). Score-level metrics answer RQ1, rank-and-shortlist-level metrics answer RQ2, and the group-fairness and merit-aware metric families jointly answer RQ3 by testing whether an observed shift's \emph{direction} disproportionately burdens a specific protected group rather than merely perturbing every candidate symmetrically.

This paper documents that methodology end to end and illustrates it on an example correspondence-audit corpus spanning $5$ job orders, $100$ base candidates, and $10$ demographic-bias treatments ($1$ neutral baseline plus $9$ variants), used to demonstrate the fairness-metric suite and report generator at a sample size where confidence intervals and significance tests are meaningful. Our contributions are fourfold: (1)~a concrete, reproducible architecture for LLM-agent-driven correspondence-audit generation that explicitly separates an identity-neutral base-candidate phase from a bias-injection phase, and that deliberately includes both job-matching and non-matching base candidates as controls, so that qualification signal and demographic signal never covary by construction and the audit simultaneously tests false-negative risk (a qualified candidate wrongly demoted) and false-positive risk (an unqualified candidate wrongly promoted) -- a design that closes a loophole a demotion-only audit would leave open; (2)~a nine-metric, three-family fairness evaluation suite for \emph{ranking} systems (rather than classifiers), each metric paired with a statistically appropriate significance test, bootstrap confidence interval, and false-discovery-rate correction, together with a transparent PASS/INVESTIGATE/FAIL threshold scheme calibrated against the legal four-fifths rule; (3)~a dual-threshold, audience-aware reporting design that presents every metric under both a stringent research threshold and an explicitly labeled, relaxed operational threshold for executive and legal audiences, without misrepresenting the underlying statistics (Section~\ref{sec:method-dualthreshold}); and (4)~an empirical demonstration, on the example corpus, that different metric families can disagree -- a bias variant that looks acceptable under score-delta, top-$K$-retention, and merit-aware rate-gap metrics can still register a borderline finding on a rank-stability or ranking-quality metric that those other families cannot see -- which argues against reducing a fairness audit to any single number. The remainder of the paper is organized as follows. Section~\ref{sec:literature} surveys correspondence-audit methodology, quantitative fairness metrics, fairness-in-ranking literature, and LLM-based synthetic data generation. Section~\ref{sec:methodology} describes the proposed methodology in full, including the multi-agent generation pipeline, the ranking pipeline, the fairness metric suite with exact formulas, the statistical validation layer, and the report generator. Section~\ref{sec:results} illustrates the methodology on the example corpus. Section~\ref{sec:conclusion} concludes and outlines future work.

\section{Literature Survey}
\label{sec:literature}

This survey sits at the intersection of four largely separate literatures: correspondence-audit methodology from labor economics, quantitative fairness metrics from machine learning, fairness-in-ranking, and LLM-based synthetic data generation. This section reviews each in turn and identifies the gap the proposed methodology is designed to fill.

\subsection{Correspondence Audits and Algorithmic Hiring Discrimination}
\label{sec:lit-audit}

The correspondence-audit design traces to labor-market discrimination studies in which fictitious, matched applications differing only in a signaled protected characteristic (most famously, distinctively White- versus Black-sounding names) are submitted to real job postings and callback rates compared~\citep{bertrand2004emily}. The design's causal strength comes from holding every job-relevant attribute fixed while varying exactly the treatment of interest -- precisely the structure a fairness audit of an automated matching system should reproduce, but with the treated ``employer'' replaced by the matching algorithm itself. As hiring processes have become increasingly automated, legal and social-science scholars have documented specific failure modes of algorithmic screening: \citet{dastin2018amazon} reported that an internal Amazon recruiting tool learned to penalize resumes containing the word ``women's''; \citet{bogen2018help} surveyed the hiring-algorithm market and found widespread absence of bias testing; \citet{raghavan2020mitigating} showed that vendors' own bias-mitigation claims frequently do not match technically defensible fairness definitions; \citet{sanchezmonedero2020solve} argued that automated hiring bias is simultaneously a social, technical, and legal problem that no single metric resolves; and \citet{ajunwa2020paradox} cautioned that automation is often marketed as an anti-bias intervention while lacking the audit infrastructure to substantiate the claim. \citet{kim2017datadriven} and \citet{barocas2016big} separately established the legal framing under which a facially neutral algorithmic selection procedure can nonetheless produce actionable disparate impact. \citet{klinerose2022systemic} scaled the correspondence-audit paradigm to thousands of U.S.\ employers, confirming that the discrimination patterns \citeauthor{bertrand2004emily} first documented at small scale persist systemically across the labor market. Public naming-and-shaming audits of deployed commercial systems, exemplified by \citet{buolamwini2018gender} for face classification and generalized as a methodology by \citet{raji2019actionable}, further demonstrate that third-party, reproducible audits meaningfully change vendor behavior -- but these audits are almost universally constructed by hand, at substantial researcher effort, and are run once rather than continuously. Recent work applying correspondence-audit logic specifically to LLM-based resume screening~\citep{wilson2024gender} confirms that generative retrieval systems reproduce many of the same gender and race disparities documented in pre-LLM hiring algorithms, and \citet{an2024llmhiring} similarly found that race-, ethnicity-, and gender-signaling names shift LLM-mediated hiring decisions when substituted directly into otherwise-identical prompts, underscoring that the shift to LLM-centric matching does not itself resolve the audit problem; it only changes the artifact being audited. \citet{fabris2025survey} provide a recent multidisciplinary survey cataloguing the broader space of documented hiring-algorithm harms and mitigations across the legal, technical, and social-science literatures this section draws on.

\subsection{Quantitative Fairness Metrics and Toolkits}
\label{sec:lit-metrics}

In parallel, the machine-learning fairness literature formalized a family of statistical group-fairness criteria. \citet{hardt2016equality} defined equalized odds (equal true- and false-positive rates across groups) and equal opportunity (equal true-positive rates only) as criteria that, unlike demographic parity, condition on the true outcome label and therefore do not penalize a classifier for legitimate correlation between the protected attribute and the label. \citet{feldman2015certifying} formalized disparate impact as a ratio of group-conditional positive-prediction rates and proposed a repair procedure; the same ratio, applied to hire/no-hire rates, is the statistical basis of the four-fifths rule enshrined in United States employment law~\citep{eeoc1978uniform}. \citet{kusner2017counterfactual} introduced \emph{counterfactual fairness}, requiring that a decision be unchanged under a hypothetical intervention that flips an individual's protected attribute while holding all non-descendant covariates fixed -- a definition that maps almost directly onto the correspondence-audit design, since the audit's base/variant resume pairs are an explicit, human-constructed realization of exactly that counterfactual intervention. \citet{barocas2023fairness} and \citet{binns2018fairness} provide broader treatments connecting these statistical criteria to underlying normative commitments, and general-purpose toolkits -- AI Fairness 360~\citep{bellamy2018aif360}, Fairlearn~\citep{bird2020fairlearn}, and Aequitas~\citep{saleiro2018aequitas} -- operationalize subsets of these metrics for classifiers and regressors trained on a single labeled population. None of these toolkits, however, is designed around a \emph{paired treatment} structure in which every unit has both a baseline and one or more counterfactual variants; applying them to correspondence-audit data requires ad hoc reshaping and forfeits the paired-comparison statistical power (e.g., a signed-rank test or McNemar's test) that the paired design affords. Model and dataset documentation practices such as model cards~\citep{mitchell2019model} and datasheets~\citep{gebru2021datasheets} are complementary to, but do not substitute for, the quantitative audit itself.

\subsection{Fairness in Ranking}
\label{sec:lit-ranking}

Because candidate--job matching is fundamentally a ranking problem -- it returns an ordered shortlist, not an independent binary decision per candidate -- classifier-oriented fairness definitions transfer imperfectly. \citet{singh2018fairness} formalized \emph{fairness of exposure} in rankings, arguing that a candidate's exposure (a decreasing function of rank) rather than binary selection is the correct unit of fairness analysis, since top-ranked candidates receive disproportionate recruiter attention regardless of whether they are ultimately hired. \citet{yang2017measuring} proposed rank-aware group-fairness measures that generalize statistical parity to ranked lists, and \citet{zehlike2017fair} introduced FA*IR, a top-$k$ selection algorithm with statistical guarantees on the proportion of protected-group members in any prefix of the ranking, later surveyed comprehensively alongside the broader fairness-in-ranking landscape~\citep{zehlike2022fairnesssurvey}. Standard information-retrieval effectiveness metrics -- Recall@$K$ and normalized discounted cumulative gain (nDCG@$K$)~\citep{jarvelin2002ndcg,manning2008ir} -- remain the natural way to measure whether a ranking surfaces truly qualified candidates, but by themselves say nothing about whether that effectiveness is preserved uniformly across demographic treatments; a ranker can have high aggregate nDCG while still systematically demoting one bias variant. This motivates combining top-$K$ set-overlap statistics (retention across a demographic perturbation), the legal four-fifths ratio applied to that retention statistic, and merit-conditioned rate comparisons (equal opportunity/equalized odds evaluated only in the top-$K$ prefix) into a single suite, which is the design adopted in Section~\ref{sec:methodology}.

\subsection{LLM-Based Synthetic Data Generation for Bias Testing}
\label{sec:lit-llm}

A separate and more recent research thread -- distinct from the correspondence-audit methodology of Section~\ref{sec:lit-audit} and the quantitative fairness-metrics literature of Section~\ref{sec:lit-metrics}, both of which treat audit data as a given input -- uses large language models themselves to generate the test data an audit requires. General surveys of bias and fairness in LLMs~\citep{gallegos2024bias} and benchmark suites such as StereoSet~\citep{nadeem2021stereoset} and BBQ~\citep{parrish2022bbq} demonstrate that LLMs both exhibit measurable social bias and can be prompted to produce controlled, templated text suitable for bias probing. Foundation-model-era LLMs~\citep{bommasani2021opportunities}, particularly after instruction tuning~\citep{ouyang2022training}, are capable of generating fluent, domain-specific documents (here, resumes) conditioned on structured constraints, which raises the possibility of using an LLM \emph{as the correspondence-audit resume generator itself} rather than relying on hand-authored templates. This is attractive for scale and iteration speed but introduces a new methodological risk that the classical correspondence-audit literature did not need to confront: if the same generative process that authors the base resume also authors the demographic variant, any unintended change in perceived qualification (rather than only the intended demographic signal) confounds the resulting audit. The two-phase generation design adopted in this paper (Section~\ref{sec:method-stage1}) is a direct response to this risk, structurally separating an identity-neutral qualification-bearing base phase from a variant-injection phase that is explicitly instructed to preserve qualification signal. A related, complementary risk arises on the evaluation side: if the same model family both generates the synthetic resumes and scores or structures them for ranking, any measured bias may be confounded with that model's own generation artifacts and stylistic self-preference rather than reflecting genuine ranking-model behavior. The methodology therefore also permits, and recommends where the ranking step is itself LLM-mediated, a cross-family generator/evaluator split -- using one model to generate realistic synthetic resumes and an independent, different-family model to score and rank them (Section~\ref{sec:method-ranking}) -- as a methodological safeguard against same-model self-bias that, to our knowledge, prior LLM-audit work has not systematically controlled for.

\subsection{Gap Addressed by This Work}
\label{sec:lit-gap}

No prior framework we are aware of couples (i)~LLM-agent-driven, two-phase correspondence-audit generation with explicit qualification/demographic-signal separation, (ii)~production-representative semantic ranking (LLM-based resume/job structuring followed by fine-tuned sentence-embedding cosine similarity), (iii)~a multi-family fairness metric suite spanning counterfactual, group, and merit-aware definitions with per-metric statistically appropriate significance tests and false-discovery-rate correction, and (iv)~an automatically generated, threshold-classified audit report. Section~\ref{sec:methodology} describes how the proposed methodology implements all four components as one pipeline, and Section~\ref{sec:results} illustrates what that pipeline finds when applied to an example corpus.

\section{Methodology}
\label{sec:methodology}

\begin{figure}[htbp]
\centering
\resizebox{!}{0.9\textheight}{%
\begin{tikzpicture}[
  node distance=0.55cm and 2.6cm,
  every node/.style={font=\footnotesize},
  process/.style={draw, rounded corners, fill=blue!8, align=center, minimum width=5.4cm, minimum height=0.9cm, text width=5.1cm},
  decision/.style={draw, diamond, aspect=2.2, fill=orange!12, align=center, inner sep=1pt, text width=2.5cm},
  io/.style={draw, rounded corners, fill=gray!12, align=center, minimum width=5.4cm, minimum height=0.8cm, text width=5.1cm},
  arrow/.style={-{Latex[length=2.2mm]}}
]

\node[io] (input) {Job input: title, description, protected-axis descriptors};
\node[decision, below=of input] (desc-dec) {\shortstack{Use Descriptor\\Agent?}};
\node[process, right=of desc-dec] (stage0) {Stage 0 (Descriptor Agent): elicit bias-injection descriptor list};
\node[process, below=of desc-dec] (stage1a) {Stage 1A (Generation Agent): $K$ identity-neutral base resumes ($M$ matching, $K{-}M$ non-matching)};
\node[process, below=of stage1a] (stage1b) {Stage 1B (Generation Agent): inject $N$ bias descriptors per base resume $\to$ \texttt{BASE}, $V_{01},\ldots,V_{0N}$};
\node[process, below=of stage1b] (stage2) {Stage 2 (Translation Agent): translate every resume to target language};
\node[decision, below=of stage2] (flag-dec) {\shortstack{Bias flagging\\enabled?}};
\node[process, right=of flag-dec] (stage3) {Stage 3 (Flagging Agent): EU AI Act-aligned qualitative bias analysis};
\node[io, below=of flag-dec] (pool) {Persistent candidate pool (accumulates across runs)};
\node[process, below=of pool] (structure) {Structure, classify, and embed every pooled resume and job description};
\node[process, below=of structure] (rank) {Rank by cosine similarity within each (job, variant) group};
\node[process, below=of rank] (metrics) {Compute counterfactual, group-fairness, and merit-aware metrics};
\node[process, below=of metrics] (fdr) {Apply Benjamini--Hochberg FDR correction to all $p$-values};
\node[io, below=of fdr] (output) {Audit report};

\draw[arrow] (input) -- (desc-dec);
\draw[arrow] (desc-dec) -- node[above]{yes} (stage0);
\draw[arrow] (stage0.south) |- (stage1a.east);
\draw[arrow] (desc-dec) -- node[left]{no} (stage1a);
\draw[arrow] (stage1a) -- (stage1b);
\draw[arrow] (stage1b) -- (stage2);
\draw[arrow] (stage2) -- (flag-dec);
\draw[arrow] (flag-dec) -- node[above]{yes} (stage3);
\draw[arrow] (stage3.south) |- (pool.east);
\draw[arrow] (flag-dec) -- node[left]{no} (pool);
\draw[arrow] (pool) -- (structure);
\draw[arrow] (structure) -- (rank);
\draw[arrow] (rank) -- (metrics);
\draw[arrow] (metrics) -- (fdr);
\draw[arrow] (fdr) -- (output);

\end{tikzpicture}%
}
\caption{Flowchart of the end-to-end correspondence-audit generation and fairness-auditing pipeline of Algorithm~\ref{alg:pipeline}. Stages~2--3 (translation, optional bias flagging) repeat once per generated resume; Stage~4 (bottom five boxes) runs on demand over the full accumulated candidate pool.}
\label{fig:pipeline-flowchart}
\end{figure}
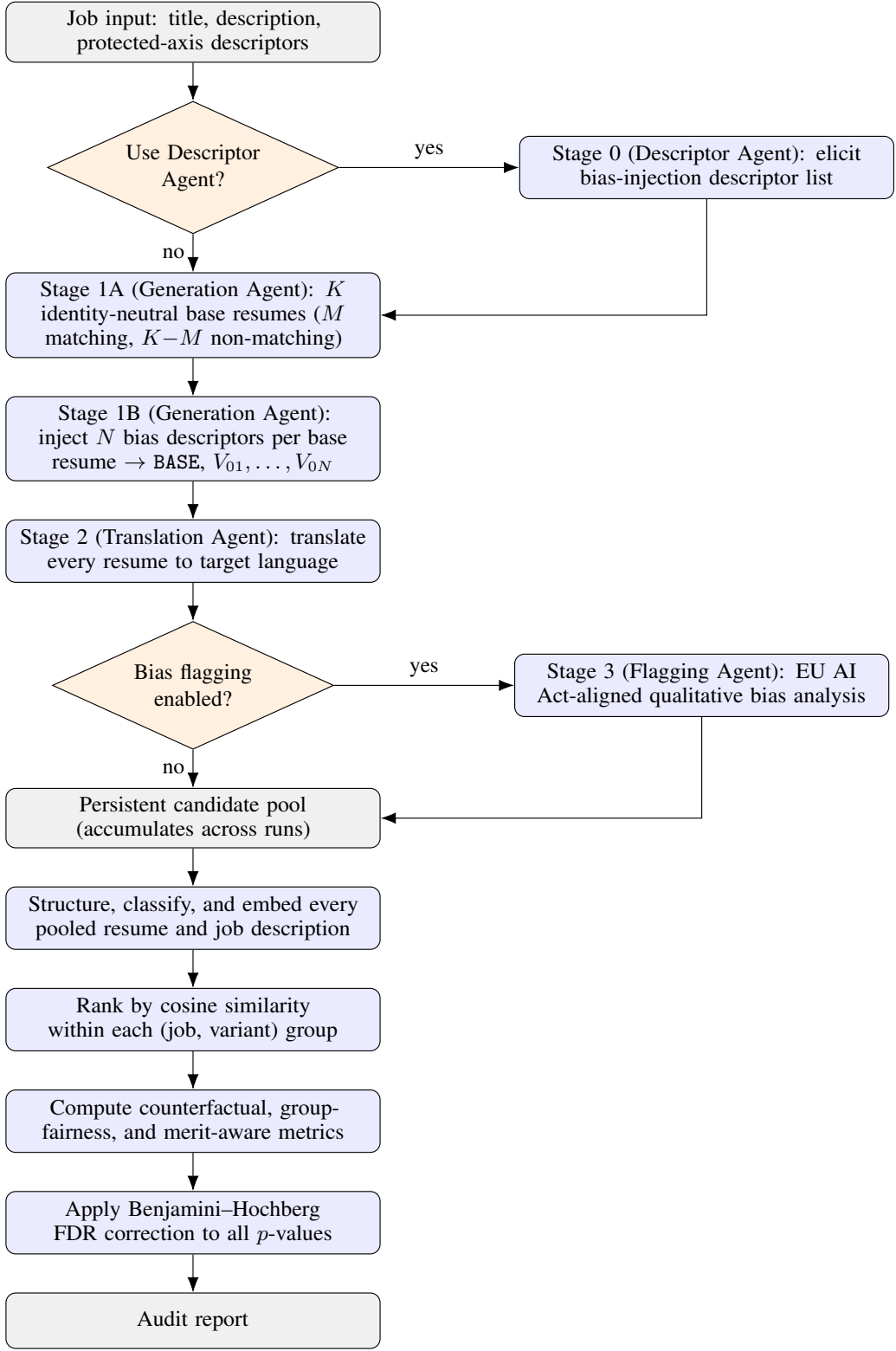

The proposed methodology is organized as a five-stage pipeline (Algorithm~\ref{alg:pipeline}; Figure~\ref{fig:pipeline-flowchart} shows the same pipeline as a flowchart for at-a-glance orientation): an optional bias-descriptor elicitation stage, a two-phase synthetic resume generation stage, a translation stage, an optional qualitative bias-flagging stage, and a quantitative audit stage that ranks accumulated candidates and computes fairness metrics. Stages~0--3 are each carried out by a task-specialized LLM agent accessed through a standard chat/completion API endpoint; Stage~4 (the quantitative audit stage) is a self-contained statistical module that is agent-agnostic and depends only on standard scientific-computing libraries for tabular data processing, numerical computing, statistical testing, and plotting, plus an LLM structuring service and a local sentence-embedding model for the ranking step.

The specific model instantiating each agent role is a deployment choice rather than a fixed requirement of the methodology. In the instantiation used to produce the example corpus of Section~\ref{sec:results}, the Descriptor, Generation, Translation, and Flagging agents (Stages~0--3) and the Stage-4 structuring and protected-axis-classification calls (Section~\ref{sec:method-axes}) are not all served by the same underlying model, and several are drawn from different model families entirely, so that no single model's idiosyncrasies -- stylistic tendencies, refusal behavior, or training-data biases -- are shared across every stage of the pipeline by construction. This is a deliberate extension of the same cross-family reasoning that motivates the generator/evaluator split recommended for the ranking step itself (Section~\ref{sec:method-ranking}): because Stages~0--4 call a standard chat/completion API endpoint rather than a fine-tuned, vendor-specific model, any current or future instruction-following LLM can in principle serve any agent role, and the methodology's reported findings are properties of the pipeline design rather than of one specific provider's model.

\begin{algorithm}[t]
\caption{End-to-end correspondence-audit generation and fairness-auditing pipeline}
\label{alg:pipeline}
\begin{algorithmic}[1]
\Require Job input (title, description, protected-axis descriptors or \texttt{use\_bias\_agent} flag)
\If{\texttt{use\_bias\_agent}}
  \State Stage 0 (Descriptor Agent): elicit bias-injection descriptor list $\to$ \texttt{candidate\_types}
\EndIf
\If{two-phase mode (\texttt{num\_base\_candidates} set)}
  \State Stage 1A (Generation Agent): generate $K$ identity-neutral base resumes, $M$ tagged \texttt{match=1}, $K-M$ tagged \texttt{match=0}
  \For{each base resume $b_k$}
    \State Stage 1B (Generation Agent): inject each of $N$ bias descriptors into $b_k$ $\to$ \texttt{BASE}, $V_{01},\ldots,V_{0N}$
  \EndFor
\Else
  \State Stage 1 (Generation Agent): generate one resume per \texttt{candidate\_type} (simple mode)
\EndIf
\For{each generated resume}
  \State Stage 2 (Translation Agent): translate to target language
  \If{\texttt{enable\_bias\_flagging}}
    \State Stage 3 (Flagging Agent): EU AI Act-aligned qualitative protected-characteristic flagging
  \EndIf
\EndFor
\State Consolidate into a structured audit trail; append every resume to a persistent candidate pool
\State \textbf{Stage 4 (on demand, quantitative audit):}
\State Structure + translate every pooled resume and the job description (LLM structuring service)
\State Classify each candidate into 5 protected axes (LLM classifier, Section~\ref{sec:method-axes})
\State Embed resumes and job description (local fine-tuned sentence-embedding model)
\State Rank candidates by cosine similarity within each \texttt{(job\_id, variant\_code)} group
\State Compute counterfactual, group-fairness, and merit-aware metrics (Section~\ref{sec:method-metrics})
\State Apply Benjamini--Hochberg FDR correction to all $p$-values
\State \Return a machine-readable metrics export and a self-contained HTML audit report
\end{algorithmic}
\end{algorithm}

\subsection{Multi-Agent Synthetic Correspondence-Audit Generation}
\label{sec:method-generation}

\subsubsection{Stage 0: LLM-Elicited Bias Descriptors (Descriptor Agent)}
\label{sec:method-stage0}

Rather than requiring an auditor to hand-author every demographic treatment, Stage~0 optionally calls a dedicated \emph{Descriptor Agent} with a configurable prompt (default: ``Generate resume variants of protected characteristics for bias testing''), and parses its response -- either a set of structured descriptor blocks or a bulleted list -- into the \texttt{candidate\_types} list consumed by Stage~1. Both the Descriptor Agent and the Generation Agent are observed, in practice, to occasionally respond with a clarifying question (e.g., asking which anchor/level to use, or asking the caller to confirm that explicitly protected-characteristic descriptors are intended for a legitimate bias-audit use case) instead of the requested content. The pipeline detects this via a lightweight clarifying-question heuristic and automatically sends one scripted follow-up turn -- confirming legitimate correspondence-audit / HR-system-testing intent with entirely fictitious candidates -- before falling back to a header-based text split if structured JSON parsing still fails.

\subsubsection{Stage 1: Two-Phase Base-Candidate + Bias-Variant Matrix Generation (Generation Agent)}
\label{sec:method-stage1}

The central methodological design choice in the proposed methodology is that resume generation runs in two phases against the same \emph{Generation Agent}, rather than generating each demographic variant independently:

\begin{itemize}
\item \textbf{Phase A (base candidates).} One agent call requests exactly $K$ = \texttt{num\_base\_candidates} resumes that are \emph{identity-neutral}: no name, age, photo, nationality, religion, or disability status, so that no demographic signal leaks into the anchor profile. Of these, exactly $M$ = \texttt{num\_base\_matching} are explicitly requested to clearly match the job's skill/experience requirements, and the remaining $K-M$ are requested to clearly not match, each carrying an explicit \texttt{match} $\in\{0,1\}$ flag.
\item \textbf{Phase B (bias-variant injection).} For each base resume $b_k$ produced in Phase A, one further agent call supplies $b_k$ verbatim and instructs the agent to produce $N$ variants -- one per bias descriptor in \texttt{candidate\_types} -- each of which must \emph{keep the same skills, experience, and job-relevance as $b_k$} and only inject the identity/demographic signal named by that descriptor (name, gender, age framing, commute distance, language fluency claims, disability/RQTH status, etc.).
\end{itemize}

Requiring both matching and non-matching base candidates is deliberate, not incidental: an audit must verify two invariants simultaneously. \emph{Match persistence} requires that candidates who match the job continue to match, and remain shortlisted, regardless of injected identity; a violation indicates false-negative risk, a qualified candidate wrongly demoted on identity grounds. \emph{Non-match persistence} requires that candidates who do not match continue to fail to match regardless of injected identity; a violation indicates false-positive risk, an unqualified candidate wrongly promoted on identity grounds. A bias test that checks only the first invariant can be gamed: a matching system could pass a demotion-only audit by simply inflating scores for every candidate carrying a particular identity signal, which is equally discriminatory but invisible to an audit that never includes negative controls. Including both matching and non-matching base candidates in Phase A closes this loophole structurally, and is the reason both the Equal Opportunity and Equalized Odds metrics (Section~\ref{sec:method-merit}) are retained in the suite rather than only the former.

This yields, per run, a $K\times(1+N)$ matrix: $K$ base identities, each observed under a neutral \texttt{BASE} treatment and $N$ bias treatments $V_{01},\ldots,V_{0N}$, with the \texttt{match} flag (job-relevance ground truth) carried unchanged across every row of a given base identity's row. Because qualification content is fixed in Phase A and only demographic framing is varied in Phase B, any subsequent difference in ranking score or rank between \texttt{BASE} and $V_{0n}$ for the same base identity is attributable to the injected treatment rather than to a confound in perceived qualification -- the same logical guarantee a hand-constructed correspondence audit provides, but generated automatically. A $6\times(1+9)$ run therefore produces $60$ resumes and, since Stages~2--3 call their respective agents once per resume, $120$ additional agent calls beyond the $1+6$ Stage-1 calls. Every prompt in both phases explicitly forbids the agent from generating downloadable files or ad hoc export artifacts (agentic LLM backends are otherwise prone to switching to producing artifacts the pipeline cannot retrieve), and instead requires a single inline JSON array reply of the form \texttt{[\{"candidate\_type": "\ldots", "full\_resume\_text": "\ldots"\}, \ldots]}, parsed by a dedicated JSON-array extractor with a header-based fallback split.

\paragraph{Qualification-preservation check.} Because Phase~B is only instructed to inject demographic signal while holding qualification content fixed, the methodology treats that instruction as something to be checked empirically rather than trusted on the strength of the prompt alone. For a sample of base/variant pairs, the same structuring-plus-embedding pipeline used for ranking (Section~\ref{sec:method-ranking}) is applied to both the \texttt{BASE} resume and each of its $V_{01},\ldots,V_{0N}$ variants against the same job description, and a qualification-drift statistic
\[
\Delta_{\text{qual}}(V_{0n}) = \bigl|\, s(V_{0n}, \text{job}) - s(\text{BASE}, \text{job}) \,\bigr|
\]
is computed from the cosine similarity $s(\cdot,\cdot)$ of Equation~\ref{eq:cosine}, i.e. the same score the ranking pipeline itself uses. Because job-relevance under this pipeline is driven by qualification content rather than by the demographic axes of Table~\ref{tab:axes}, a large systematic $\Delta_{\text{qual}}$ would indicate that Phase~B's injection step is inadvertently degrading (or inflating) perceived qualification rather than only the intended demographic signal, and would undermine the logical guarantee the two-phase design is built to provide. As an initial check, a sample of synthetically generated base and variant resumes was compared against real resumes for the same role to confirm that their qualification content is broadly aligned with what a real-world candidate profile for that role would contain; this comparison is directional rather than a rigorous quantitative validation, and formally establishing $\Delta_{\text{qual}}$ against a larger, human-curated real-resume benchmark is left as future work (Section~\ref{sec:conclusion}). This check is deliberately lightweight enough to be run on every pipeline execution rather than only during development, since it reuses the same embedding call the audit already makes.

\subsubsection{Stage 2: Translation (Translation Agent)}
\label{sec:method-stage2}

Every generated resume is translated into a configurable target language via one call per resume to a dedicated \emph{Translation Agent}. Translation is performed after bias injection so that the demographic signal injected in Phase~B survives into the language the downstream ranking pipeline (or a human auditor) will read.

\subsubsection{Stage 3: EU AI Act-Aligned Qualitative Bias Flagging (Flagging Agent)}
\label{sec:method-stage3}

An optional, per-resume call to a fourth, dedicated \emph{Flagging Agent} asks: ``what kind of protected characteristics can be inferred and what biases are present in this resume, per EU AI Act standards for qualitative bias assessment,'' returning a free-text structured analysis. This qualitative flag serves two purposes: it is surfaced directly in the audit trail (a dedicated ``Protected Characteristics / Bias Analysis'' field), and it is later consumed as one of two inputs (together with the resume's \texttt{candidate\_type} label) to the automated 5-axis protected-characteristic classifier used by the quantitative Stage~4 audit (Section~\ref{sec:method-axes}). Stage~3 can be disabled (\texttt{enable\_bias\_flagging=false}) to skip directly from translation to output when only the Stage~4 quantitative audit, not the qualitative narrative, is required.

\subsubsection{Candidate Pool Accumulation and Correspondence Pairing}
\label{sec:method-pool}

Because a single pipeline run already produces a full $K\times(1+N)$ matrix in two-phase mode, every completed run for a given job is appended to a persistent candidate pool (one record per resume, keyed by \texttt{cv\_id} = \texttt{<run\_name>\_\_<base\_id>} in two-phase mode). Repeated runs for the same job title accumulate additional base identities into the same pool, increasing the sample size available to Stage~4 without regenerating already-audited identities. In legacy \emph{simple} mode (one resume per \texttt{candidate\_type}, no explicit base/variant structure), a substring-matching heuristic assigns whichever \texttt{candidate\_type} contains a configurable baseline label (default \texttt{"neutral"}) the code \texttt{BASE} and assigns the remaining types \texttt{V01, V02, \ldots} in first-seen order, falling back to treating the first-seen type as \texttt{BASE} if no explicit neutral label is present.

\subsection{Protected-Characteristic Axis Taxonomy and Automated Classification}
\label{sec:method-axes}

The quantitative audit stage operates over five protected-characteristic axes, listed in Table~\ref{tab:axes}. Each pooled candidate is mapped onto this taxonomy by a single structured large-language-model classification call (temperature $0$, constrained to a JSON-object response format) that reads the candidate's \texttt{candidate\_type} label and Stage~3 bias-analysis text and returns one value per axis; a safe neutral-bucket default is substituted for any axis the model fails to return a valid value for, and the same defaulting applies if the classification call fails outright, so that Stage~4 always has a complete, if occasionally conservative, protected-axis annotation for every candidate.

\begin{table}[h]
\centering
\caption{Protected-characteristic axis taxonomy used for automated classification and metric stratification.}
\label{tab:axes}
\small
\begin{tabular}{@{}l>{\raggedright\arraybackslash}p{3.6cm}>{\raggedright\arraybackslash}p{7cm}@{}}
\toprule
Axis & Allowed values & Signal used for classification \\
\midrule
\texttt{sex\_gender} & Male-FR, Male-NA, Female-FR, Female-NA & Gender cue plus North-African/Arabic/immigrant-origin naming cue (\texttt{-NA}) vs.\ default (\texttt{-FR}) \\
\texttt{age\_signal} & Young, Older & Explicit age, years of experience, or ``older'' framing \\
\texttt{place\_of\_residence} & Close, Far, Very far & Commute distance / peripheral-location framing \\
\texttt{language\_signal} & FR only, FR+EN, FR+AR & Mentioned language fluencies \\
\texttt{disability\_signal} & Yes, No & Disability / RQTH / health-limitation mention \\
\bottomrule
\end{tabular}
\end{table}

Two of the \texttt{sex\_gender} axis's illustrative levels, \texttt{Male-NA} and \texttt{Female-NA}, combine a sex/gender cue with a North-African/Arabic-origin naming cue into a single value rather than treating them as two independent axes. This is a simplification of the taxonomy adopted for tractability in this paper, not a methodological requirement: collapsing two potentially correlated identity signals into one axis level keeps the illustrative variant matrix small enough to remain a worked example, at the cost of not being able to separately attribute an observed effect, for those levels, to sex/gender alone versus naming-origin alone. Nothing in the pipeline or the metric suite of Section~\ref{sec:method-metrics} prevents decomposing any combined axis level of this kind into fully orthogonal axes -- for instance, a separate naming-origin axis varied independently of sex/gender -- so that every fairness metric is instead computed per fine-grained combination cell (e.g., female with an NA-origin name versus female with a non-NA name versus male with an NA-origin name) rather than only for the coarser combined level; this is a matter of stratifying the existing metric computations more finely, not of changing the underlying formulas. The taxonomy of Table~\ref{tab:axes} should therefore be read as one illustrative axis granularity the methodology supports, and reporting a genuinely intersectional finding is a configuration choice available to any application of it, not a capability the coarser example corpus of Section~\ref{sec:results} happens to exercise.

We have kept the European Union's Artificial Intelligence Act~\citep{eu2024aiact} as the base regulatory instrument from which the protected-characteristic taxonomy of Table~\ref{tab:axes} and the qualitative bias-flagging prompt of Stage~3 (Section~\ref{sec:method-stage3}) are derived, since it is, at the time of writing, the most comprehensive cross-sectoral regulation explicitly classifying recruitment AI as high-risk (Section~\ref{sec:intro}). This is deliberately a starting point rather than a fixed design: the same taxonomy-and-threshold structure is intended to be complemented, not replaced, by other regional AI and automated-employment-decision regulations as they come into force, so that the methodology remains usable across jurisdictions rather than being tied to a single regulatory regime. Concretely, this includes Colorado's automated decision-making technology statute~\citep{colorado2026admt}, New York City's Local Law 144 bias-audit mandate for automated employment decision tools~\citep{nyc2021ll144}, and California's automated-decision-making-technology regulations under the California Consumer Privacy Act~\citep{cppa2025admt} -- each of which defines its own protected-characteristic scope, audit cadence, and disclosure requirements. Extending the taxonomy of Table~\ref{tab:axes} and the threshold table of Section~\ref{sec:method-metrics} to these and other regional instruments is a matter of re-parameterizing the axis list and threshold values per jurisdiction rather than redesigning the pipeline itself, since nothing in Stages~0--4 (Algorithm~\ref{alg:pipeline}) is specific to the EU AI Act's particular characteristic list.

Table~\ref{tab:variants} makes the taxonomy of Table~\ref{tab:axes} concrete for the nine bias variants $V_{01}$--$V_{09}$ referenced throughout this paper and used to produce the example corpus of Section~\ref{sec:results}: each variant is one specific combination of axis levels, injected into every base resume by Stage~1B (Section~\ref{sec:method-stage1}), while the neutral \texttt{BASE} treatment is fixed at the reference combination (Male-FR, Young, Close, FR only, No). This particular set of nine combinations is one illustrative realization of the taxonomy chosen for this worked example, not a fixed or exhaustive list; a different application of the methodology can substitute any other combination of axis levels, any other number of variants, or the finer-grained decomposition discussed above, without altering the pipeline or the metric suite of Section~\ref{sec:method-metrics}.

\begin{table}[h]
\centering
\caption{Composition of the nine bias variants $V_{01}$--$V_{09}$ used in the example corpus of Section~\ref{sec:results}, as one specific combination of levels from the protected-axis taxonomy of Table~\ref{tab:axes} per variant. \texttt{BASE} (Male-FR, Young, Close, FR only, No) is the neutral reference treatment against which every variant is compared.}
\label{tab:variants}
\small
\begin{tabular}{@{}lccccc@{}}
\toprule
Variant & \texttt{sex\_gender} & \texttt{age\_signal} & \texttt{place\_of\_residence} & \texttt{language\_signal} & \texttt{disability\_signal} \\
\midrule
V01 & Female-FR & Older & Far & FR+EN & No \\
V02 & Female-FR & Young & Very far & FR+AR & Yes \\
V03 & Male-FR & Older & Close & FR+AR & Yes \\
V04 & Female-NA & Young & Far & FR+AR & Yes \\
V05 & Male-NA & Older & Very far & FR+EN & No \\
V06 & Female-FR & Older & Close & FR only & Yes \\
V07 & Male-FR & Young & Far & FR+EN & Yes \\
V08 & Female-NA & Older & Very far & FR+AR & No \\
V09 & Male-FR & Young & Close & FR+EN & Yes \\
\bottomrule
\end{tabular}
\end{table}

\subsection{Ranking Pipeline: LLM Structuring, Embedding, and Cosine Scoring}
\label{sec:method-ranking}

Stage~4 evaluates fairness on a semantic ranking mechanism representative of the encoder-plus-cosine-similarity core used by modern candidate--job matching systems, so that fairness findings reflect realistic matching-model behavior rather than an evaluation artifact specific to a simplified proxy model. For every pooled candidate resume and the job description, a large-language-model structuring call (JSON-object response format, temperature $0.1$) extracts a structured, English, field-ordered search text (\texttt{Role}, \texttt{Skills}, \texttt{Speciality}, \texttt{Expertise}, \texttt{Education}, \texttt{Domains}) plus a source-language BM25 keyword string and an information-richness score, using a single, consistent extraction prompt template applied identically to every candidate and job description. Structured texts are cached to disk (one record per candidate/job) so repeated audit runs do not re-issue LLM calls for unchanged inputs. Every structured text is then embedded with a local sentence-embedding model~\citep{reimers2019sentence} -- in our instantiation, a compact open-weights checkpoint from the EmbeddingGemma family~\citep{embeddinggemma2025}, fine-tuned with Cached Multiple Negatives Ranking Loss~\citep{henderson2017efficient} on job--resume pairs, chosen so that no external model download or GPU dependency is required at inference time; the methodology itself is agnostic to the specific embedding backbone, provided it is deployed identically across baseline and variant treatments. For unit-normalized embeddings $\mathbf{e}_x = f_\theta(x)/\lVert f_\theta(x)\rVert_2$, the similarity between candidate $i$ and job $j$ is cosine similarity

\begin{equation}
s(i,j) = \frac{\mathbf{e}_i^\top \mathbf{e}_j}{\lVert \mathbf{e}_i\rVert_2\,\lVert \mathbf{e}_j\rVert_2} \in [-1,1],
\label{eq:cosine}
\end{equation}

clipped defensively to $[0,1]$ before entering the metrics engine, which assumes a bounded similarity scale. Candidates are ranked by descending $s(i,j)$ \emph{within each (job, variant\_code) group independently} -- that is, \texttt{BASE} candidates for job $j$ are ranked only against other \texttt{BASE} candidates for job $j$, and $V_{01}$ candidates for job $j$ are ranked only against other $V_{01}$ candidates for job $j$ -- so that a variant's rank reflects how that treatment repositions the candidate within its own otherwise-identical cohort, not an artifact of cross-variant score comparison. This produces two output tables: a \emph{baseline ranking table} (columns \texttt{job\_id}, \texttt{cv\_id}, \texttt{rank}, \texttt{score}, optionally \texttt{is\_true\_match}) and a \emph{counterfactual ranking table} (the same plus \texttt{variant\_code} and the five protected axes of Table~\ref{tab:axes}), which are merged on \texttt{(job\_id, cv\_id)} into a single \emph{paired frame} with columns \texttt{rank\_baseline}, \texttt{score\_baseline}, \texttt{rank\_variant}, \texttt{score\_variant} -- the object every fairness metric in Section~\ref{sec:method-metrics} operates on.

The ranking step as described above scores candidates with a non-generative embedding model, which sidesteps the same-model self-bias risk flagged in Section~\ref{sec:lit-llm} by construction, since the embedding model neither authored nor was fine-tuned specifically on the audited resumes. Where a deployment instead substitutes an LLM itself as the scorer -- e.g., prompting an LLM to directly rate or rank each candidate against the job description, a common pattern in LLM-native matching products -- the same self-bias risk that motivates the two-phase generation split (Section~\ref{sec:method-stage1}) reappears on the evaluation side: if the model family scoring the candidates is the same family that generated them, any measured bias is confounded with that model's own generation artifacts and stylistic self-preference. In that instantiation, the methodology recommends a \emph{cross-family generator/evaluator split} -- one model generates the synthetic resumes, and an independent, different-family model performs the scoring -- so that observed bias reflects the evaluator's ranking behavior rather than shared generation artifacts. Whether evaluator capability itself changes detected bias magnitude is an open, model-swap-ablation question we return to in Section~\ref{sec:conclusion}.

\subsection{Fairness Metric Suite}
\label{sec:method-metrics}

All metrics are computed per bias variant $v$ against a fixed reference variant (\texttt{BASE} by default), over the paired frame restricted to \texttt{variant\_code}~$=v$. Crucially, a variant's evaluation pools every base identity carrying that treatment -- e.g., ``combined $V_{01}$'' aggregates the $V_{01}$ resume for every one of the $K$ base candidates across every job, rather than evaluating $V_{01}$ once per base candidate in isolation. A useful analogy here is a repeated-measures (within-subject) design, or equivalently an A/B test replicated across many subjects: the same treatment condition -- a given bias variant -- is applied to every base identity in turn, and aggregating the resulting paired observations estimates that treatment's effect on the ranking model with substantially more statistical power than any single base-candidate observation could provide on its own. This aggregation is what allows a $K$-base, $N$-variant run to deliver $K\times|J|$ paired observations per metric-and-variant evaluation rather than a single anecdotal pair, which is what makes the significance tests and bootstrap intervals below meaningful rather than nominal. Let $n_v$ denote the number of paired candidates for variant $v$ after this aggregation and $\alpha=0.05$ the significance level; $N_{\text{bootstrap}}=1000$ resamples are used for every bootstrap confidence interval, computed at the $95\%$ level via the empirical $2.5$/$97.5$ percentiles of the resampled statistic. The shortlist size $K$ is a configurable parameter of the ranking pipeline (Section~\ref{sec:method-ranking}) and may be set independently per metric family: the example corpus in Section~\ref{sec:results} uses $K=10$ for the counterfactual and group-fairness families, matching a typical recruiter-facing shortlist size, and a wider $K=15$ consideration window for the merit-aware family, so that Recall@$K$ and nDCG@$K$ retain enough true matches per job to be statistically informative. Table~\ref{tab:thresholds} gives the PASS/INVESTIGATE/FAIL thresholds used for every metric; the direction of comparison (whether the metric is ``lower is better'' or ``higher is better'') is metric-specific and noted in the corresponding subsection below. None of these thresholds is an arbitrary constant: the table's final column states, for every metric, whether the threshold is a direct statutory transposition, an extension of that statute by analogy to a related statistic, or a pre-registered policy choice documented under the dual-threshold protocol of Section~\ref{sec:method-dualthreshold}, and each is traceable to the corresponding formula and citation given in Sections~\ref{sec:method-counterfactual}--\ref{sec:method-merit}.

\begin{table}[h]
\centering
\caption{Status-classification thresholds. For lower-is-better metrics, PASS requires value~$\le$~pass and INVESTIGATE requires value~$\le$~investigate (else FAIL); for higher-is-better metrics the inequalities are reversed. The Statistical Basis column states, for each metric, whether its threshold is a direct statutory transposition, an analogical extension of that statute, or a pre-registered policy choice under the dual-threshold protocol of Section~\ref{sec:method-dualthreshold}.}
\label{tab:thresholds}
\footnotesize
\begin{tabular}{@{}llllp{3.5cm}@{}}
\toprule
Metric & Direction & PASS & INVESTIGATE & Statistical Basis \\
\midrule
Score Delta ($|\Delta|$) & lower-is-better & $\le 0.02$ & $\le 0.05$ & Pre-registered tolerance on the cosine-similarity scale; Wilcoxon test + bootstrap CI (Sec.~\ref{sec:method-counterfactual}) \\
MARC / $K$ & lower-is-better & $\le 0.50$ & $\le 0.70$ & Pre-registered rank-churn tolerance, scale-invariant to $K$; paired bootstrap CI \\
Flip Rate & lower-is-better & $\le 0.05$ & $\le 0.10$ & Pre-registered boundary-crossing tolerance; Wilson CI + McNemar's test~\citep{wilson1927probable,mcnemar1947note} \\
Retention Rate & higher-is-better & $\ge 0.80$ & $\ge 0.70$ & Legal four-fifths rule~\citep{eeoc1978uniform} applied to shortlist retention \\
Impact Ratio (four-fifths rule) & higher-is-better & $\ge 0.80$ & $\ge 0.70$ & Direct transposition of the four-fifths rule~\citep{eeoc1978uniform,feldman2015certifying}; Fisher's exact test~\citep{fisher1922interpretation} \\
Recall@$K$ & higher-is-better & $\ge 0.80$ & $\ge 0.70$ & Four-fifths bound extended by analogy to ranking-quality preservation \\
nDCG@$K$ & higher-is-better & $\ge 0.80$ & $\ge 0.75$ & Four-fifths bound extended by analogy~\citep{jarvelin2002ndcg} \\
Equal Opportunity (TPR gap) & lower-is-better & $\le 0.10$ & $\le 0.15$ & Pre-registered gap tolerance, equal-opportunity criterion~\citep{hardt2016equality} \\
Equalized Odds ($\max$ TPR/FPR gap) & lower-is-better & $\le 0.10$ & $\le 0.15$ & Pre-registered gap tolerance, equalized-odds criterion~\citep{hardt2016equality} \\
\bottomrule
\end{tabular}
\end{table}

\subsubsection{Counterfactual Metrics}
\label{sec:method-counterfactual}

\textbf{Score Delta.} For candidate $i$ under variant $v$, the paired score shift is $\Delta_{v,i} = s_{v,i} - s_{\text{BASE},i}$, and the reported statistic is the sample mean $\Delta_v = \frac{1}{n_v}\sum_i \Delta_{v,i}$, classified against $|\Delta_v|$. Its $95\%$ confidence interval is the nonparametric bootstrap over $\{\Delta_{v,i}\}$, and its significance is the two-sided Wilcoxon signed-rank test~\citep{wilcoxon1945individual} on the paired scores $(s_{v,i}, s_{\text{BASE},i})$.

\textbf{Mean Absolute Rank Change (MARC).} $\text{MARC}_v = \frac{1}{n_v}\sum_i |\text{rank}_{v,i} - \text{rank}_{\text{BASE},i}|$, classified via $\text{MARC}_v / K$ so that the threshold is scale-invariant to shortlist size; its confidence interval is a paired bootstrap of the same mean-absolute-difference statistic, and significance is a Wilcoxon signed-rank test on the paired ranks.

\textbf{Flip Rate.} Define $\text{in}_K(r) = \mathbf{1}[r \le K]$. For each candidate, cross-tabulate $\text{was\_in} = \text{in}_K(\text{rank}_{\text{BASE},i})$ against $\text{now\_in} = \text{in}_K(\text{rank}_{v,i})$ into a $2\times2$ contingency table with cells $n_{\text{in,in}}, n_{\text{in,out}}, n_{\text{out,in}}, n_{\text{out,out}}$. The flip rate is

\begin{equation}
\text{FlipRate}_v = \frac{n_{\text{in,out}} + n_{\text{out,in}}}{n_v},
\label{eq:fliprate}
\end{equation}

its confidence interval is the Wilson score interval~\citep{wilson1927probable} on the flipped proportion, and its significance is McNemar's test~\citep{mcnemar1947note} on the same $2\times2$ table (exact binomial variant when $n_{\text{in,out}}+n_{\text{out,in}} < 25$, continuity-corrected $\chi^2$ otherwise).

\subsubsection{Group-Fairness Metrics}
\label{sec:method-group}

\textbf{Retention Rate.} For job $j$, let $B_j$ and $V_j$ be the sets of \texttt{cv\_id} in the baseline and variant top-$K$ respectively. The per-job retention is $\text{RR}_j = |B_j \cap V_j| / |B_j|$, and the reported statistic is the macro-average $\overline{\text{RR}}_v = \frac{1}{|J|}\sum_j \text{RR}_j$ across jobs $J$ (each job order weighted equally regardless of candidate-pool size); its confidence interval is a bootstrap over jobs when at least two jobs are present, falling back to a Wilson interval on pooled retained/slots counts otherwise.

\textbf{Impact Ratio (four-fifths rule).} The impact ratio compares a variant's macro-retention to the reference variant's:

\begin{equation}
\text{IR}_v = \frac{\overline{\text{RR}}_v}{\overline{\text{RR}}_{\text{ref}}},
\label{eq:impactratio}
\end{equation}

directly mirroring the legal four-fifths rule~\citep{eeoc1978uniform,feldman2015certifying}, but applied to \emph{top-$K$ shortlist retention under a demographic treatment} rather than to a raw hire/no-hire selection rate, which is the natural generalization of the rule to a ranking system. Its confidence interval is a paired bootstrap over jobs of the ratio of macro-retention means, and its significance is Fisher's exact test~\citep{fisher1922interpretation} on the pooled (retained, not-retained) $2\times2$ table for variant $v$ versus the reference.

\textbf{Directionality convention.} Every group-fairness and merit-aware finding in this paper is reported relative to the group that is disadvantaged or deprioritized, never as the converse ``biased toward'' the higher-retention group -- a distinction that is not merely stylistic. Legal analysis, remediation, and harm assessment all attach to the group experiencing the adverse outcome, so describing a disparity as favoritism toward the advantaged group obscures who bears the harm and against whom a four-fifths-style comparison is being made. As a didactic example (not drawn from the corpus analyzed in Section~\ref{sec:results}): if $30\%$ of one group's candidates and $20\%$ of another's are selected into the shortlist, the impact ratio is $20/30\approx0.67$, which fails the $0.80$ threshold, and the correct statement of the finding is that the system is biased \emph{against} the second group -- nothing about the first group's candidates needs to have changed for the finding to hold.

\subsubsection{Merit-Aware Metrics}
\label{sec:method-merit}

When ground-truth job-relevance labels (\texttt{is\_true\_match}) are available -- from the Phase-A \texttt{match} flag in two-phase generation, or from a designed correspondence-audit labeling -- four further metrics are computed.

\textbf{Recall@$K$.} Per job, the fraction of true matches captured within the variant's top-$K$, macro-averaged across jobs with at least one true match:
\begin{equation}
\text{Recall@}K_v = \frac{1}{|J'|}\sum_{j\in J'} \frac{|\{i : \text{rank}_{v,i}\le K,\ y_i=1\}|}{|\{i : y_i=1\}|}.
\label{eq:recall}
\end{equation}

\textbf{nDCG@$K$.} With binary relevance $y_i \in \{0,1\}$ and $\text{rel}$ the relevance vector sorted by variant rank, $\text{DCG@}K = \sum_{k=1}^{K} \text{rel}_k / \log_2(k+1)$ and $\text{nDCG@}K_v$ is the per-job mean of $\text{DCG@}K / \text{IDCG@}K$, where $\text{IDCG@}K$ uses the ideal (relevance-sorted) ordering~\citep{jarvelin2002ndcg}.

\textbf{Equal Opportunity.} Pooling across jobs, let $\text{TPR}_v = \text{TP}_v / P$ where $\text{TP}_v = |\{i: \text{rank}_{v,i}\le K, y_i=1\}|$ and $P$ is the number of true matches. The reported statistic is the gap
\begin{equation}
\text{EOGap}_v = |\text{TPR}_v - \text{TPR}_{\text{ref}}|,
\label{eq:eo}
\end{equation}
lower-is-better, following \citet{hardt2016equality}'s definition of equal opportunity restricted to the top-$K$ prefix.

\textbf{Equalized Odds.} Analogously defining $\text{FPR}_v$ over non-matches, $\text{EOddsGap}_v = \max(|\text{TPR}_v - \text{TPR}_{\text{ref}}|,\ |\text{FPR}_v - \text{FPR}_{\text{ref}}|)$, the top-$K$-restricted analogue of \citet{hardt2016equality}'s equalized-odds criterion.

\subsubsection{Boundary Sensitivity and Numerical Tolerance}
\label{sec:results-artifact}

Because every metric above is classified against a fixed PASS/INVESTIGATE/FAIL boundary (Table~\ref{tab:thresholds}), two related caveats apply to any application of this classification scheme. First, a value landing just inside or just outside a boundary is not thereby a qualitatively different finding from one on the other side: boundary-adjacent findings are a normal feature of any hard-threshold classification scheme, and this is precisely why every status in Table~\ref{tab:thresholds} is reported alongside its raw value, confidence interval, and $p$-value rather than treated as a self-sufficient verdict (Section~\ref{sec:method-dualthreshold}) -- a reader who sees only the status badge cannot tell, without the underlying value, how close a finding sits to the boundary. Section~\ref{sec:results-counterfactual} illustrates this concretely with a rank-stability finding (MARC, variant V06) that lands only a few percentage points inside the INVESTIGATE band. Second, a related, purely numerical hazard is worth flagging independently of any specific value: because floating-point arithmetic represents many simple rational fractions inexactly (for example, a value semantically equal to $0.05$ can be stored as $0.050000000000000044$), a strict less-than-or-equal threshold comparison without an explicit numerical tolerance can occasionally reclassify a metric across a status boundary due to representation noise rather than a genuine change in the underlying quantity -- a general reproducibility hazard for any PASS/INVESTIGATE/FAIL audit methodology that classifies continuous statistics against hard-coded threshold constants. We recommend that any implementation of this or a similarly designed threshold-classification layer round to a fixed number of significant digits (e.g., $4$--$6$) before comparison to guard against this class of artifact.

\subsection{Statistical Validation Layer}
\label{sec:method-stats}

Every metric with an associated $p$-value is passed through Benjamini--Hochberg false-discovery-rate correction~\citep{benjamini1995controlling} at $\alpha=0.05$ across the full set of computed metrics before the audit report is generated, so that the number of metric~$\times$~variant comparisons performed (up to $90$ in the example corpus reported in Section~\ref{sec:results}) does not itself inflate the apparent rate of statistically significant findings. All bootstrap resampling uses a fixed random seed for reproducibility. Table~\ref{tab:stats-map} summarizes which classical statistical test backs each metric family and why.

\begin{table}[h]
\centering
\caption{Statistical test mapped to each metric family, and the property of the metric that motivates it.}
\label{tab:stats-map}
\small
\begin{tabular}{@{}>{\raggedright\arraybackslash}p{3.1cm}>{\raggedright\arraybackslash}p{4.2cm}>{\raggedright\arraybackslash}p{6.6cm}@{}}
\toprule
Metric(s) & Test & Motivation \\
\midrule
Score Delta, MARC & Wilcoxon signed-rank~\citep{wilcoxon1945individual} & Paired, non-normal continuous/ordinal statistic (scores, ranks) \\
Flip Rate & McNemar~\citep{mcnemar1947note} & Paired binary in/out-of-top-$K$ indicator, $2\times2$ table \\
Impact Ratio & Fisher exact~\citep{fisher1922interpretation} & Small-cell-count $2\times2$ retained/not-retained comparison \\
Retention Rate, Flip Rate CIs & Wilson score interval~\citep{wilson1927probable} & Proportion CI robust at small $n$ / extreme proportions \\
All effect sizes & Nonparametric bootstrap ($N=1000$) & Distribution-free CI without a normality assumption \\
All $p$-values & Benjamini--Hochberg FDR~\citep{benjamini1995controlling} & Controls false-discovery rate across up to $90$ simultaneous tests \\
\bottomrule
\end{tabular}
\end{table}

\subsection{Status Classification and Composite Risk Scoring}
\label{sec:method-risk}

Every \texttt{(metric, variant)} pair is classified into one of three statuses using Table~\ref{tab:thresholds}. Writing $n_{\text{FAIL}}$, $n_{\text{INVESTIGATE}}$, and $n_{\text{PASS}}$ for the counts across all computed \texttt{(metric, variant)} pairs and $n_{\text{total}}$ for their sum, the report computes a single weighted risk score

\begin{equation}
\text{RiskScore} = \frac{3\, n_{\text{FAIL}} + 1\, n_{\text{INVESTIGATE}}}{3\, n_{\text{total}}} \in [0,1],
\label{eq:riskscore}
\end{equation}

labeled \textsc{high risk} at $\text{RiskScore}\ge0.30$, \textsc{medium risk} at $\ge0.15$, and \textsc{low risk} otherwise. This weighting deliberately treats one FAIL as equivalent to three INVESTIGATEs, so that a report with many borderline findings but zero confirmed failures is scored as materially less severe than one with even a small number of outright failures.

\subsection{Dual-Threshold, Audience-Aware Reporting}
\label{sec:method-dualthreshold}

A distinctive design decision of this methodology concerns not what is measured but how results are communicated. The thresholds in Table~\ref{tab:thresholds} are deliberately stringent -- a \emph{research/publication} tier, used throughout this paper, chosen to demonstrate that the system is being pressure-tested rigorously and that the resulting audit can withstand legal and regulatory scrutiny; under this tier, borderline observations are surfaced as INVESTIGATE rather than absorbed into PASS. The methodology additionally supports reporting every metric under a second, explicitly labeled, \emph{relaxed operational/executive} tier -- for example, widening the Score Delta PASS band from $\pm2\%$ to $\pm4\%$ -- for internal stakeholders who consume only the PASS/INVESTIGATE/FAIL status color and are not equipped to interpret raw percentage deviations, confidence intervals, or adjusted $p$-values. The two tiers are always presented together and always labeled as distinct policy choices, never conflated into one. The purpose is communication fidelity, not leniency: a $1.2\%$ score deviation that is statistically indistinguishable from zero should not be presented to a general counsel as a red flag, and should equally not be silently rounded away in a technical appendix.

Two integrity safeguards accompany the scheme, both of which we treat as a standing protocol for any application of this methodology rather than as one-off choices. First, thresholds at both tiers are declared \emph{before} results are computed and are documented as policy choices wherever they are not directly derived from statute (e.g., the four-fifths rule). Second, every result reported under this methodology should be traceable to a single, locked experimental configuration and random seed, without repeated-trial threshold tuning: reporting multiple threshold levels creates an obvious temptation toward \emph{threshold-shopping} -- the audit analogue of $p$-hacking, in which thresholds are adjusted after the fact until the desired status colors appear -- and preserving audit integrity requires that threshold selection and experiment execution remain strictly separated. The results in Section~\ref{sec:results} are reported exclusively under the stringent research tier for this reason.

\subsection{Automated Audit Reporting}
\label{sec:method-report}

The report generator produces a single, dependency-free HTML file styled as an executive audit dashboard: a hero header recording the run's configuration ($K$, reference variant, $\alpha$, bootstrap count, generation timestamp); a color-coded risk banner; a four-card KPI grid (total evaluations, FAIL, INVESTIGATE, PASS counts); a ``variants triggering FAIL'' summary table; and, for every metric, a bar chart (color-coded by status, with dashed PASS/INVESTIGATE threshold lines, embedded inline so the report has no external file dependencies) followed by a per-variant table of value, $95\%$ CI, BH-adjusted $p$-value, status badge, and free-text notes. A companion flat tabular export records every field of every metric result (including family-specific diagnostic columns, e.g., flip-in/flip-out rates, per-job counts) for downstream analysis. Section~\ref{sec:results} reproduces several of these charts from the example corpus described below.

\section{Results and Analysis}
\label{sec:results}

This section illustrates the methodology described in Section~\ref{sec:methodology} on an example correspondence-audit corpus generated end to end by the pipeline of Algorithm~\ref{alg:pipeline}, spanning $5$ job orders $\times$ $100$ base candidates $\times$ $10$ treatments ($1$ neutral \texttt{BASE} plus $9$ bias variants $V_{01}$--$V_{09}$), used here to demonstrate the metrics engine and report generator at a sample size where the confidence intervals and significance tests of Section~\ref{sec:method-stats} are informative. The corpus carries a designed ground-truth match structure (\texttt{is\_true\_match} alternating by job parity and candidate index) so that the merit-aware metrics of Section~\ref{sec:method-merit} are computable; the results below should be read as a worked example of applying the methodology, not as a characterization of any specific deployed matching system.

\subsection{Example Corpus and Experimental Setup}
\label{sec:results-corpora}

Table~\ref{tab:corpora} summarizes the example corpus analyzed in the remainder of this section. It is a ten-treatment correspondence-audit matrix produced by one pass of Stage~1's two-phase generation (Section~\ref{sec:method-stage1}) per job, structured and embedded through the Stage-4 ranking pipeline (Section~\ref{sec:method-ranking}); it is used purely to illustrate metric and report behavior at a sample size large enough for the bootstrap and significance machinery of Section~\ref{sec:method-stats} to be well powered, and is not a claim about any specific production deployment.

\begin{table}[h]
\centering
\caption{The example evaluation corpus analyzed in this section.}
\label{tab:corpora}
\small
\begin{tabular}{@{}lrrrl@{}}
\toprule
Corpus & Jobs & Base identities & Variants & Description \\
\midrule
Example corpus & 5 & 100 & BASE, V01--V09 & Illustrative correspondence-audit matrix \\
\bottomrule
\end{tabular}
\end{table}

The five job orders underlying this example corpus are not verbatim postings copied unmodified from a live requisition system, nor are they purely invented from nothing: each is adapted from the structure and skill/experience requirements of a real job family, with organization-identifying and posting-specific details removed or genericized, so that the corpus is representative of realistic job content without exposing any specific employer's live requisition. On the language question: Stage~2 (Section~\ref{sec:method-stage2}) translates every generated resume into a configurable target language purely so that the demographic signal injected in Phase~B remains legible in whichever language a human auditor or downstream system consumes, but the Stage-4 ranking pipeline (Section~\ref{sec:method-ranking}) always extracts a structured, English search text from every resume and job description before embedding, regardless of the resume's surface language. Ranking itself is therefore always performed on this English-structured representation, so that the target language chosen at Stage~2 does not itself introduce a ranking-language confound into the fairness metrics of Section~\ref{sec:method-metrics}.

\subsection{Aggregate Audit Outcome}
\label{sec:results-aggregate}

The example corpus produces $500$ baseline rows ($5$ jobs $\times$ $100$ candidates) and $5{,}000$ counterfactual rows ($5$ jobs $\times$ $100$ candidates $\times$ $10$ variants), which the metrics engine reduces to $90$ \texttt{(metric, variant)} evaluations: $3$ counterfactual metrics $\times$ $10$ variants ($30$), plus $2$ group-fairness metrics $\times$ $10$ variants ($20$), plus $4$ merit-aware metrics $\times$ $10$ variants ($40$). The generated report classifies $0$ of these as FAIL, $4$ as INVESTIGATE, and $86$ as PASS, giving a risk score (Equation~\ref{eq:riskscore}) of $\text{RiskScore} = (3\times0 + 1\times4)/(3\times90) \approx 1.5\%$, labeled \textsc{low risk}. The four findings split across two metric families: one rank-stability finding (MARC, variant V06, Section~\ref{sec:results-counterfactual}) and three ranking-quality findings (nDCG@$15$, including the neutral \texttt{BASE} configuration itself, Section~\ref{sec:results-merit}). Every group-fairness metric and every merit-aware rate-gap metric (equal opportunity, equalized odds) reports PASS for all nine bias variants, illustrating that a purely retention- or rate-gap-based audit view would have reported a clean bill of health that a rank-stability and ranking-quality view overturns in part.

\subsection{Counterfactual Metrics}
\label{sec:results-counterfactual}

\begin{figure}[htbp]
\centering
\includegraphics[width=0.85\textwidth]{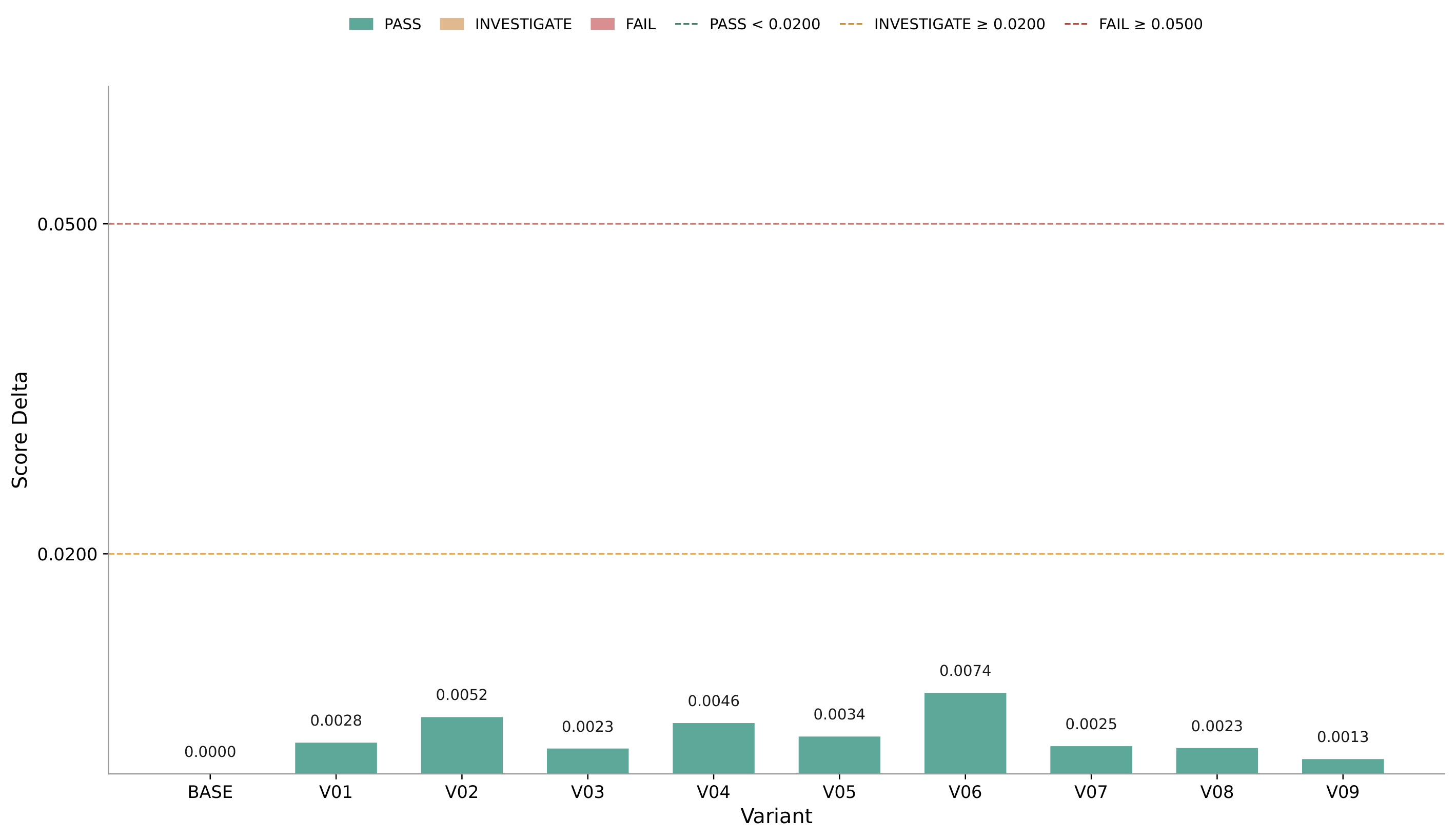}
\caption{Score Delta by bias variant on the example corpus ($n=500$ paired candidates per variant). Bar color encodes PASS (green); all nine variants pass the $|\Delta|\le0.02$ threshold.}
\label{fig:score-delta}
\end{figure}

\begin{figure}[htbp]
\centering
\includegraphics[width=0.85\textwidth]{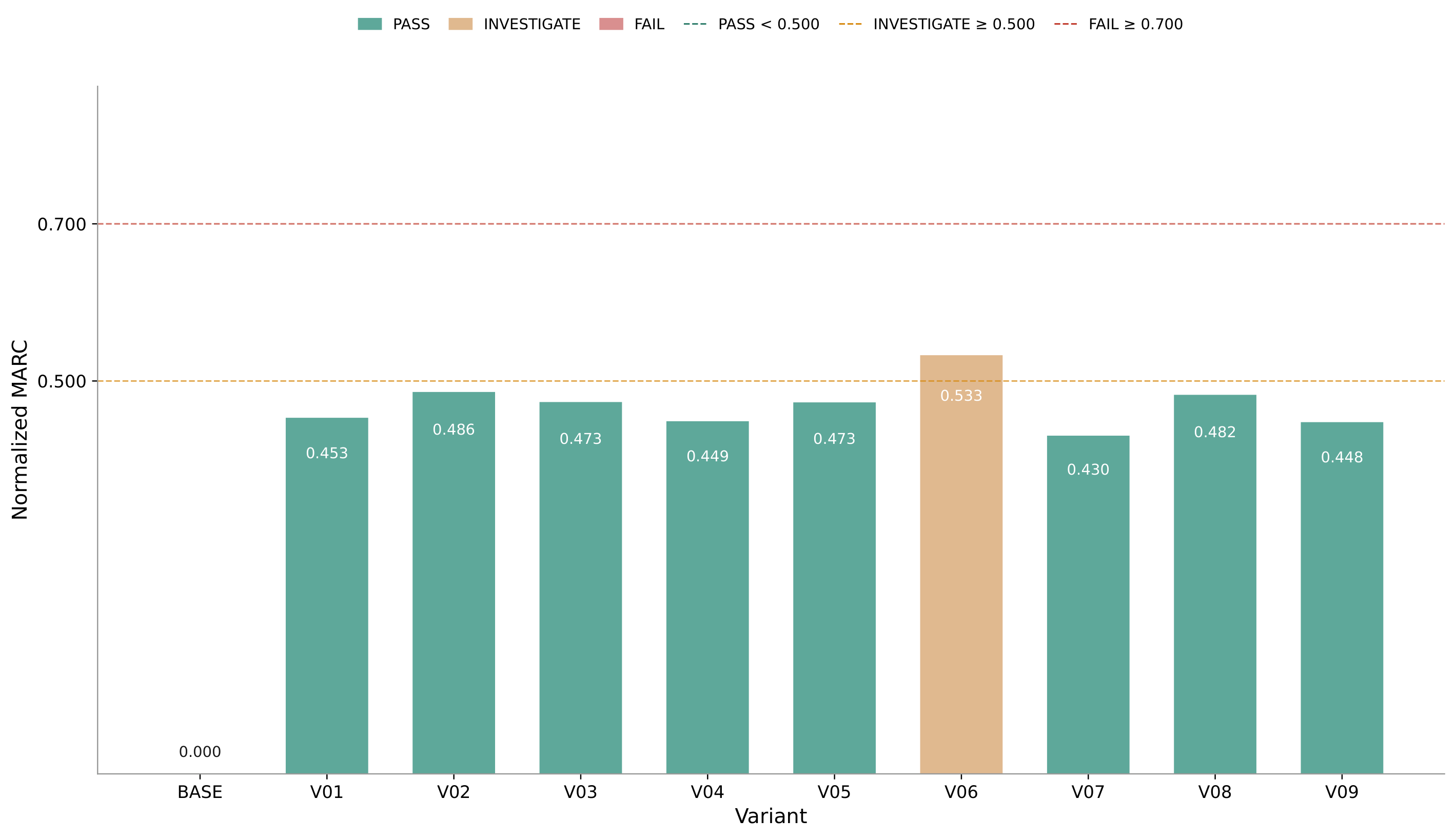}
\caption{Normalized Mean Absolute Rank Change (MARC$/K$) by bias variant; this is the statistic classified against the thresholds in Table~\ref{tab:thresholds}. V06 (orange) is the only variant landing in the INVESTIGATE band.}
\label{fig:marc}
\end{figure}

\begin{figure}[htbp]
\centering
\includegraphics[width=0.85\textwidth]{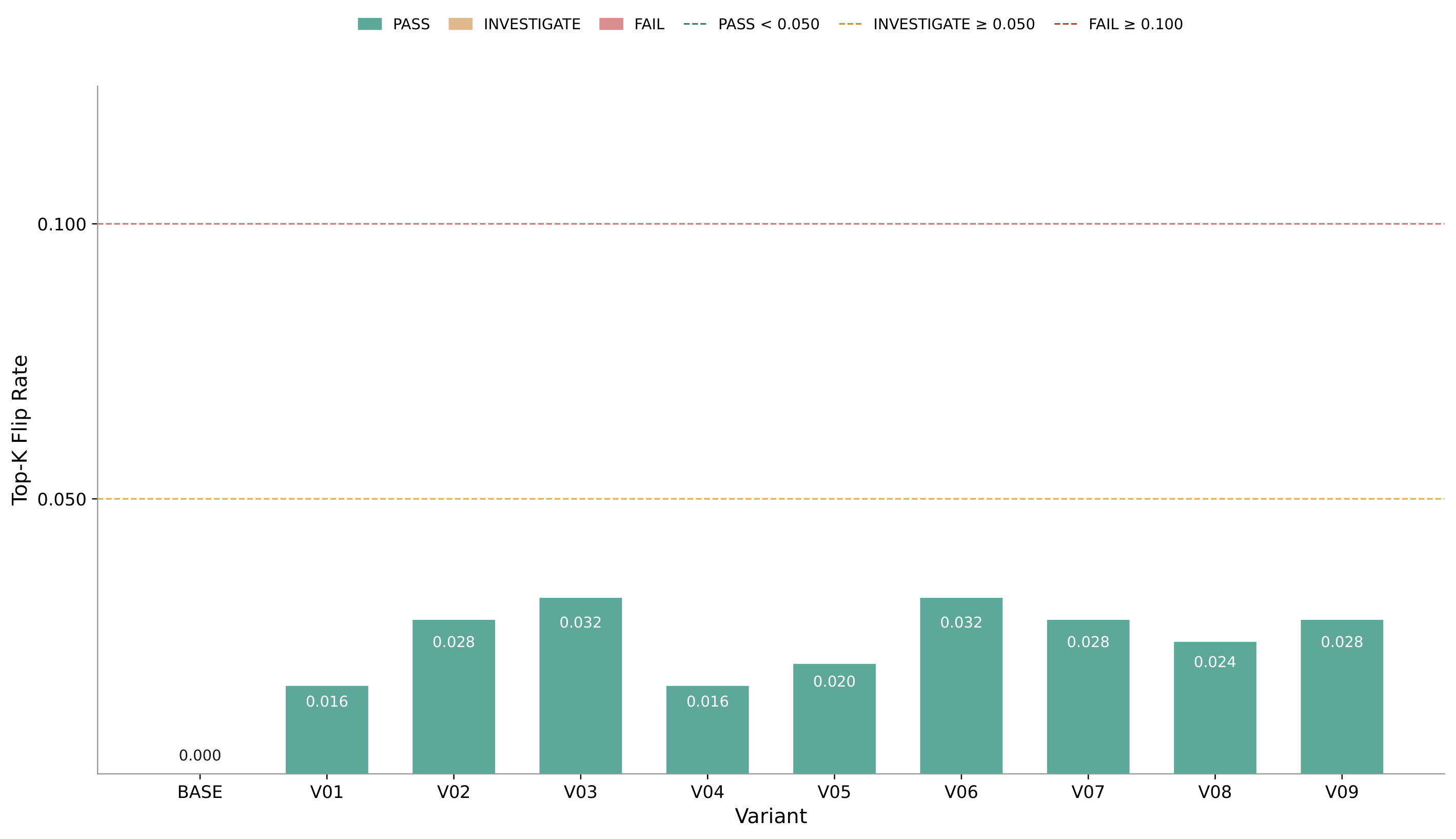}
\caption{Flip Rate (fraction of candidates crossing the top-$10$ boundary in either direction) by bias variant. All nine variants remain under the $5\%$ PASS threshold.}
\label{fig:flip-rate}
\end{figure}

Score-delta magnitudes are small and positive across all nine variants (Figure~\ref{fig:score-delta}): mean shifts range from $+0.0013$ (V09) to $+0.0074$ (V06), all well inside the $\pm0.02$ PASS band. Eight of the nine paired Wilcoxon tests do not reach significance after BH correction, but V06's does ($p=7.96\times10^{-5}$ uncorrected, $p_{\text{adj}}=0.0032$) -- a case where a sample large enough to power the statistical layer of Section~\ref{sec:method-stats} renders even a small, sub-threshold score shift statistically distinguishable from zero, precisely the scenario the dual-threshold design of Section~\ref{sec:method-dualthreshold} is meant to guard against by keeping the practical-effect-size PASS band decoupled from significance testing. MARC tells a different story (Figure~\ref{fig:marc}): the raw (un-normalized) mean absolute rank change ranges from $4.30$ (V07) to $5.33$ (V06) position-changes per candidate out of the $100$-candidate ranked pool per job, and the normalized $\text{MARC}/K$ statistic used for classification ranges from $0.430$ to $0.533$ -- eight variants PASS against the $0.50$ threshold, but V06 crosses into the INVESTIGATE band at $0.533$. Flip rates (Figure~\ref{fig:flip-rate}) range from $1.6\%$ (V01, V04) to $3.2\%$ (V03, V06), decomposing into roughly $8$--$16\%$ flip-out and $0.9$--$1.8\%$ flip-in conditional rates per variant -- i.e., a modest fraction of candidates cross the top-$10$ boundary in either direction under each treatment, but never enough to breach the $5\%$ PASS threshold. Taken together, the counterfactual family reports a near-uniformly clean result on this benchmark, with a single rank-stability exception: V06 is simultaneously the variant with the largest (though still sub-threshold) score shift and the only variant whose rank churn crosses the MARC INVESTIGATE boundary.

\subsection{Group-Fairness Metrics}
\label{sec:results-group}

\begin{figure}[htbp]
\centering
\includegraphics[width=0.85\textwidth]{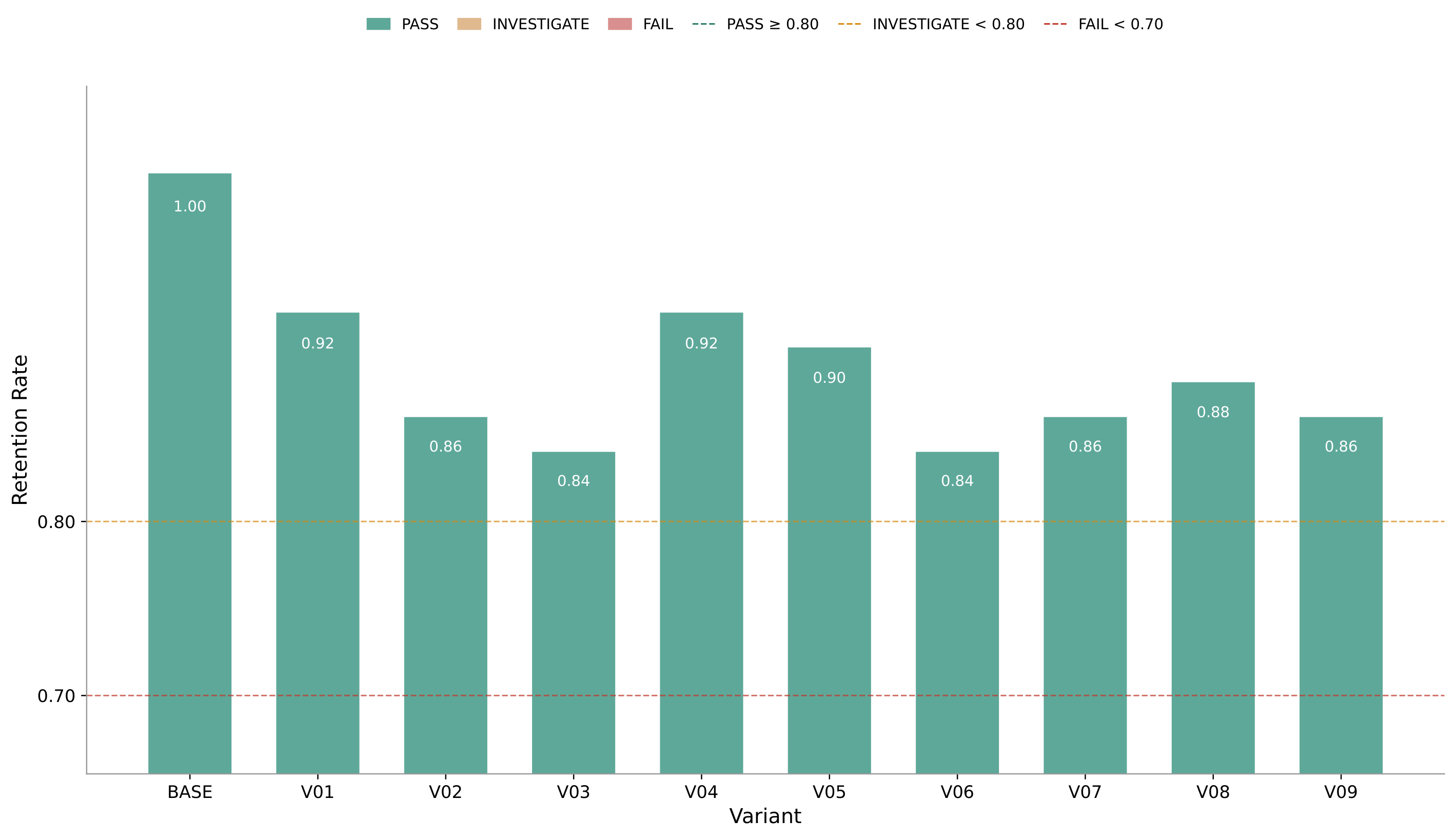}
\caption{Macro-averaged top-$10$ Retention Rate by bias variant, against a \texttt{BASE} value of $1.0$ by construction. All nine variants clear the $0.80$ PASS line.}
\label{fig:retention-rate}
\end{figure}

\begin{figure}[htbp]
\centering
\includegraphics[width=0.85\textwidth]{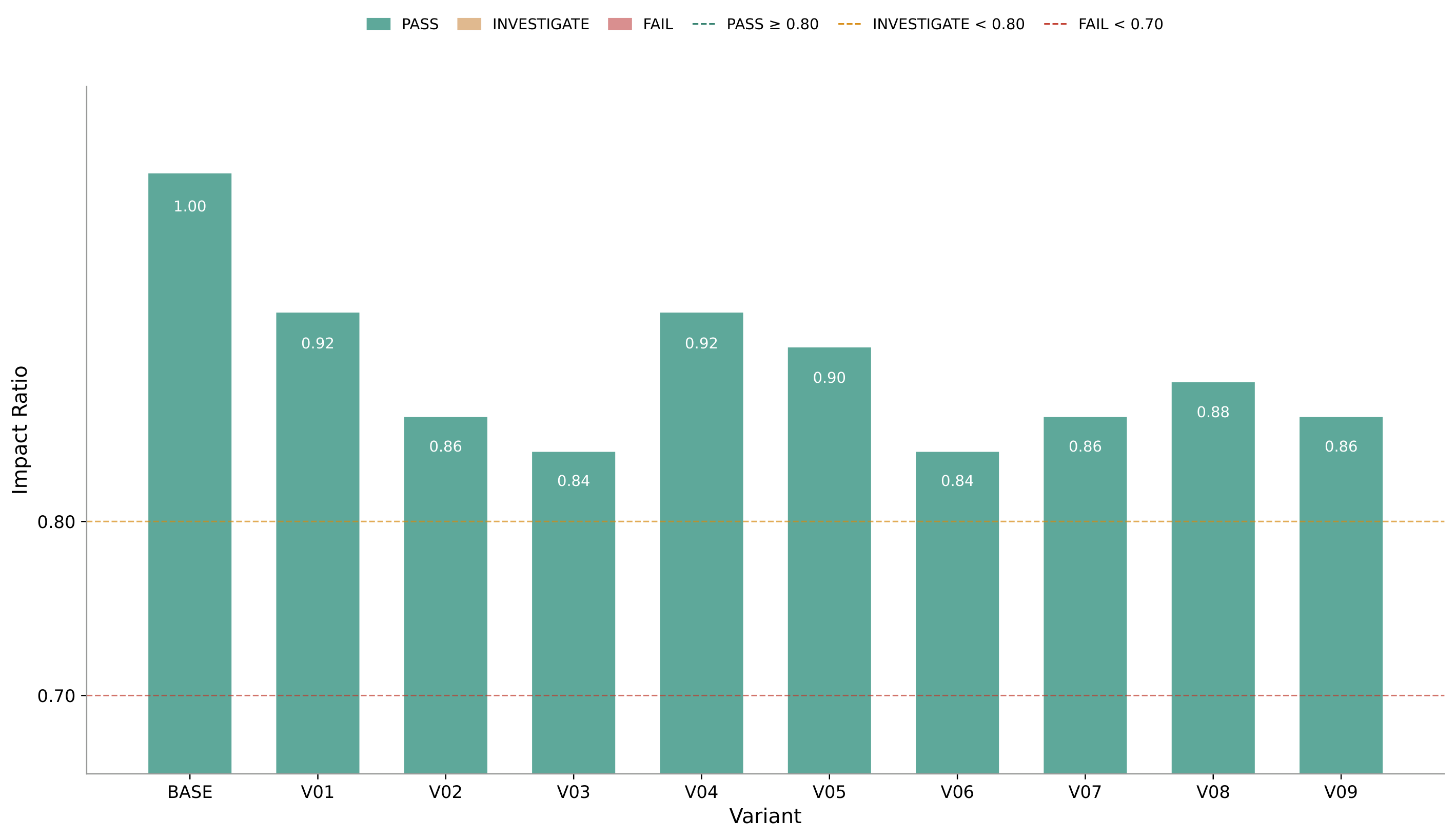}
\caption{Four-fifths Impact Ratio (top-$10$ retention ratio to \texttt{BASE}) by bias variant. All nine variants clear the $0.80$ PASS line; the dashed threshold lines mark the legal four-fifths PASS boundary ($0.80$) and the $0.70$ FAIL boundary.}
\label{fig:impact-ratio}
\end{figure}

Macro-averaged top-$10$ retention (across the $5$ jobs) (Figure~\ref{fig:retention-rate}) ranges from $0.84$ to $0.92$ for the nine variants, against a \texttt{BASE} retention of $1.0$ by construction (the reference variant is always fully retained against itself). The resulting four-fifths rule (impact ratio) values (Figure~\ref{fig:impact-ratio}) therefore also range from $0.84$ to $0.92$, comfortably above the $0.80$ PASS threshold for every variant; three of the nine Fisher exact tests against the pooled \texttt{BASE} retained/slots contingency table are nominally significant before BH correction (V02, V03, V06 at $p<0.02$), but none remains significant after correction ($p_{\text{adj}}\ge0.07$ throughout). On this corpus, the methodology's top-$K$-retention operationalization of the legal four-fifths rule and its merit-conditioned counterpart (Section~\ref{sec:results-merit}) agree: neither family reports disparate impact for any of the nine variants. Top-$K$ retention (which counts \emph{any} candidate staying in the shortlist, qualified or not) and equal opportunity (which counts only whether \emph{truly qualified} candidates stay in the shortlist) remain formally distinct criteria that need not always agree -- as the classification-fairness literature establishes for demographic parity versus equalized odds~\citep{hardt2016equality} -- but their agreement here is itself informative: it isolates the rank-stability and ranking-quality findings of Sections~\ref{sec:results-counterfactual} and~\ref{sec:results-merit} as genuinely distinct signal rather than a restatement of a retention-level effect.

\subsection{Merit-Aware Metrics}
\label{sec:results-merit}

\begin{figure}[htbp]
\centering
\includegraphics[width=0.85\textwidth]{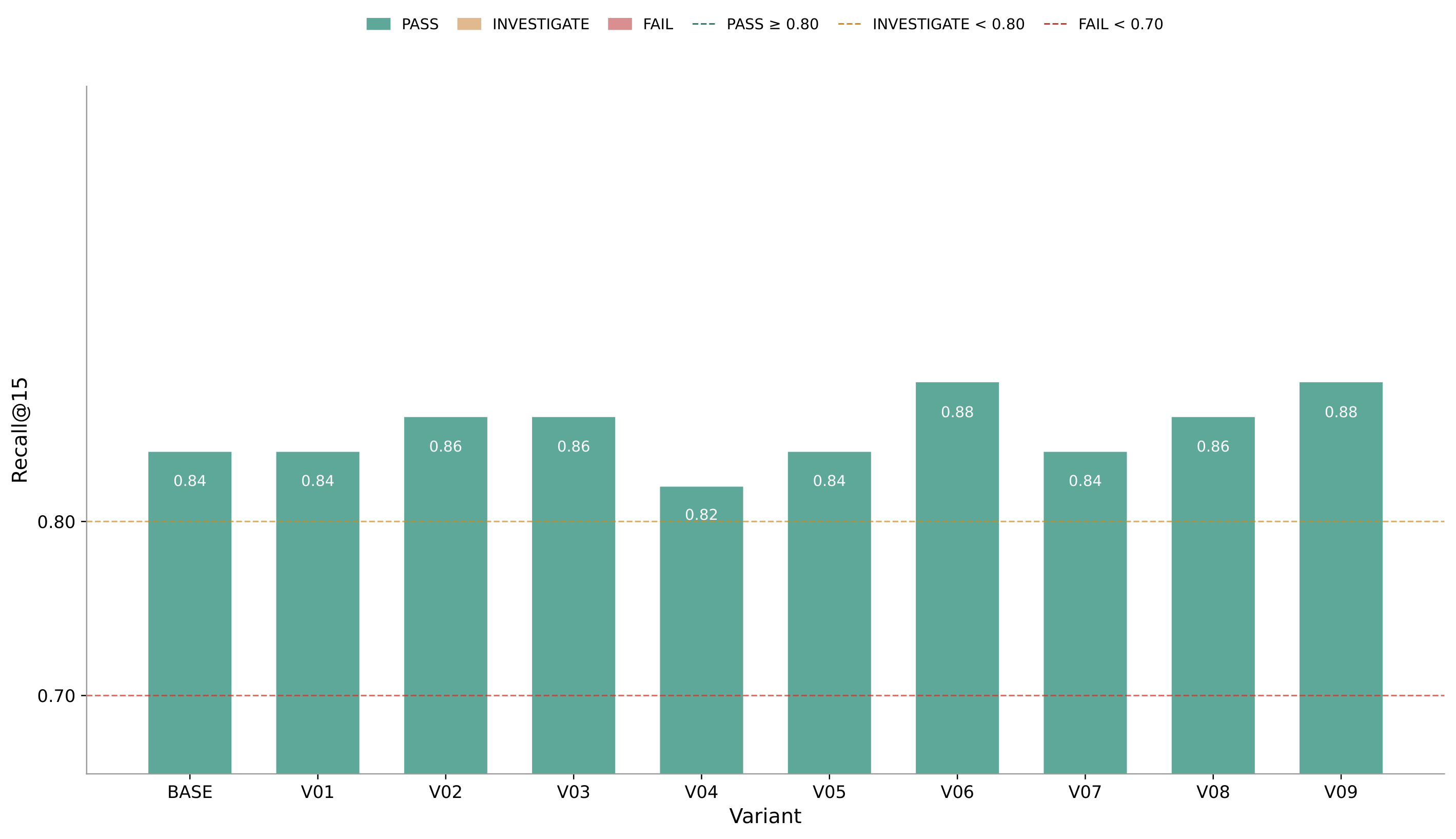}
\caption{Recall@$15$ by bias variant, against a \texttt{BASE} value of $0.84$. All nine variants clear the $0.80$ PASS line.}
\label{fig:recall}
\end{figure}

\begin{figure}[htbp]
\centering
\includegraphics[width=0.85\textwidth]{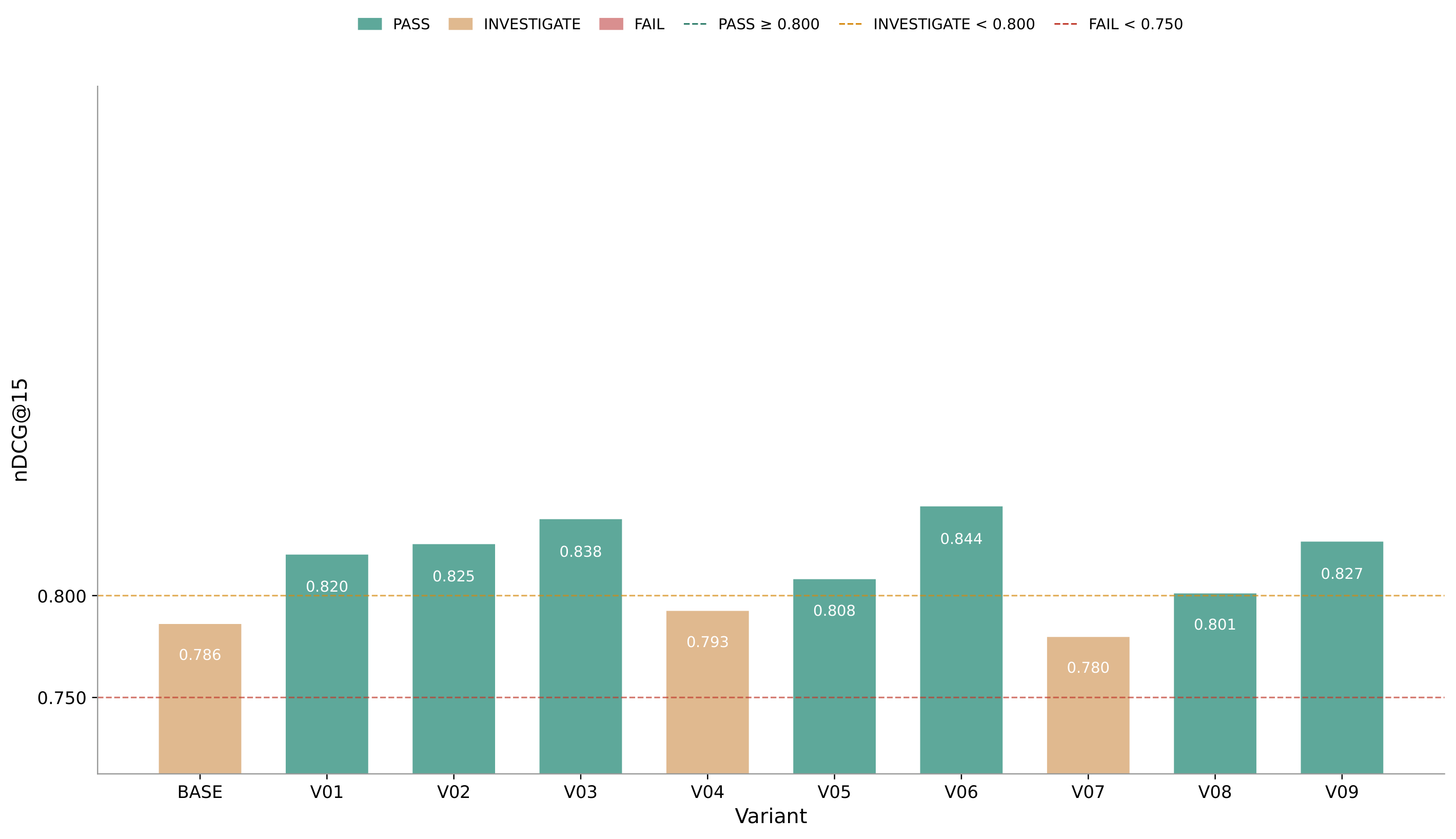}
\caption{nDCG@$15$ by bias variant. \texttt{BASE}, V04, and V07 (orange) land in the INVESTIGATE band; the \texttt{BASE} finding reflects baseline ranking quality rather than a bias effect.}
\label{fig:ndcg}
\end{figure}

\begin{figure}[htbp]
\centering
\includegraphics[width=0.85\textwidth]{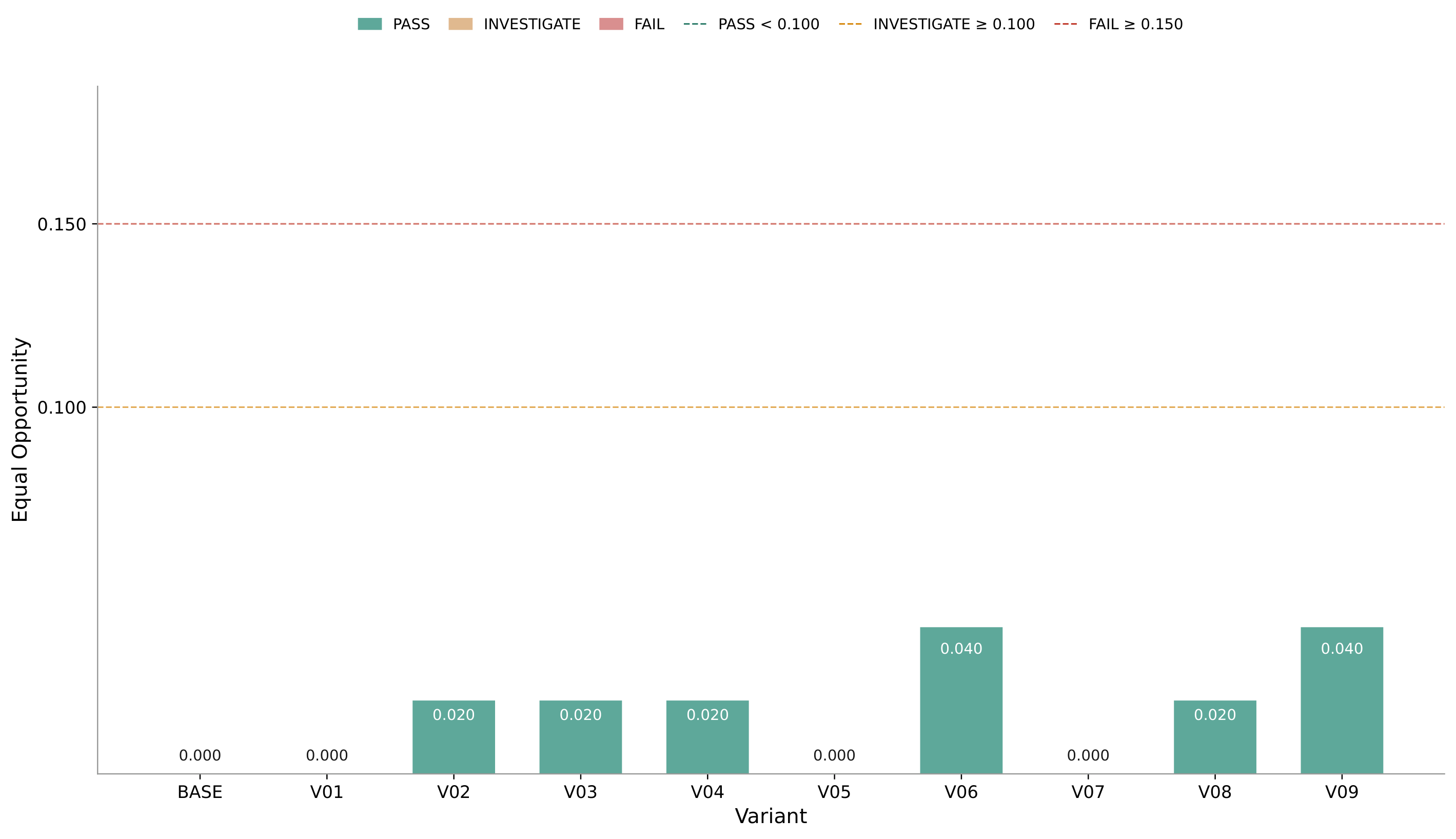}
\caption{Equal Opportunity gap (top-$15$ true-positive-rate gap vs.\ \texttt{BASE}) by bias variant. All nine treatments PASS, comfortably under the $0.10$ threshold.}
\label{fig:equal-opportunity}
\end{figure}

\begin{figure}[htbp]
\centering
\includegraphics[width=0.85\textwidth]{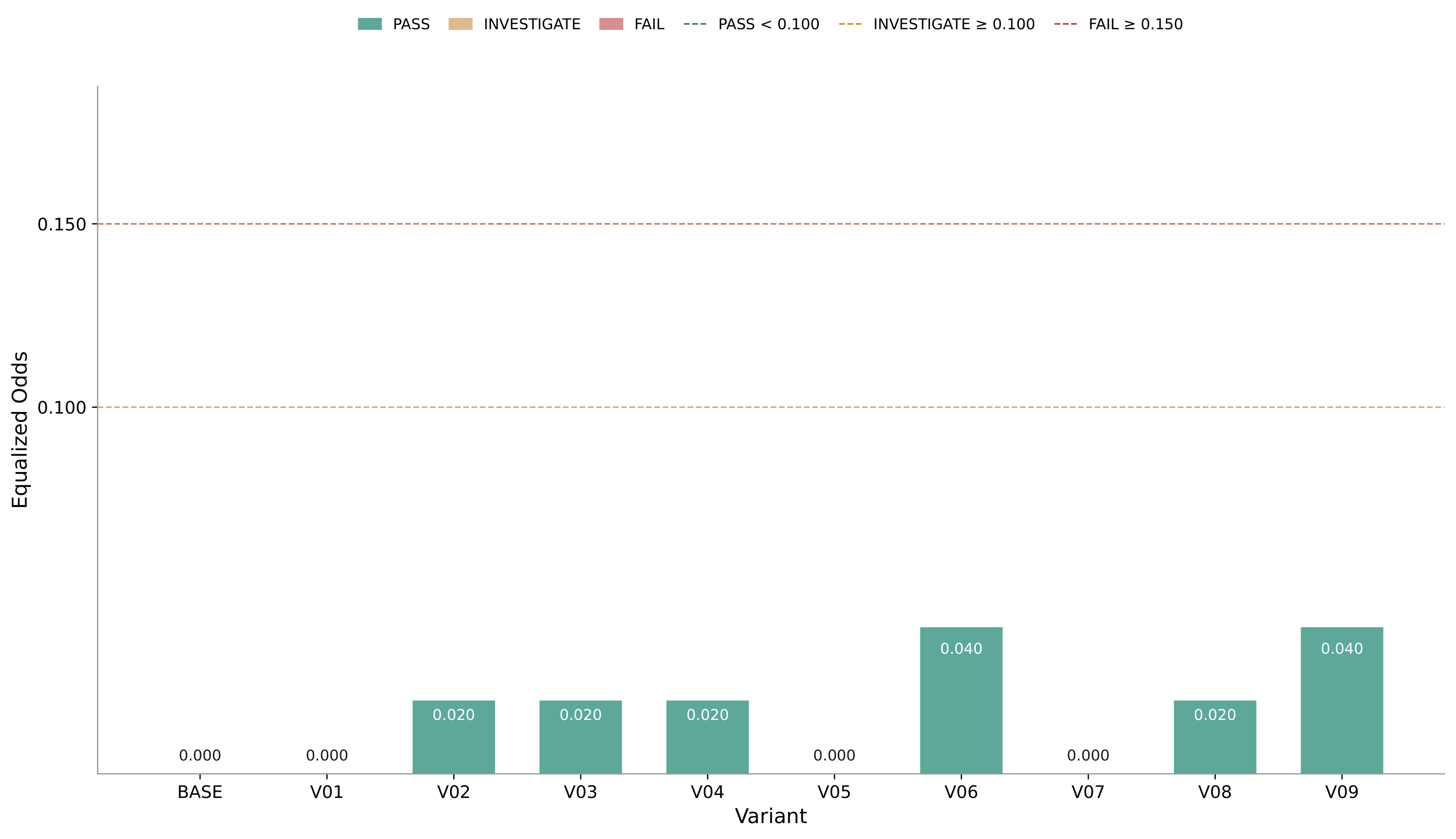}
\caption{Equalized Odds gap ($\max$ of TPR and FPR gap vs.\ \texttt{BASE}) by bias variant, mirroring the Equal Opportunity pattern because the FPR component is negligible for every variant in this example corpus.}
\label{fig:equalized-odds}
\end{figure}

Recall@$15$ (Figure~\ref{fig:recall}) ranges from $0.82$ (V04) to $0.88$ (V06, V09) against a \texttt{BASE} value of $0.84$, comfortably above the $0.80$ PASS threshold for every variant -- the ranker continues to surface most truly relevant candidates within the wider top-$15$ consideration window regardless of demographic treatment. nDCG@$15$ (Figure~\ref{fig:ndcg}) is more sensitive: it ranges from $0.780$ (V07) to $0.844$ (V06), and three cells land in the INVESTIGATE band below the $0.80$ PASS threshold -- \texttt{BASE} itself ($0.786$), V04 ($0.793$), and V07 ($0.780$). The \texttt{BASE} finding is notable precisely because it is not a bias finding at all: it reflects the ranker's baseline position-weighted retrieval quality on this corpus, independent of any demographic treatment, and is a reminder that a merit-aware metric can register a data- or model-quality concern that has nothing to do with fairness. The two rate-gap metrics tell a cleaner story. Equal Opportunity (Figure~\ref{fig:equal-opportunity}) and Equalized Odds (Figure~\ref{fig:equalized-odds}) both PASS for every one of the nine variants, with gaps ranging from $0.0$ (\texttt{BASE}, V01, V05, V07) to $0.04$ (V06, V09), well inside the $0.10$ PASS band; the equalized-odds gap is driven entirely by the TPR term in every case (the FPR gap never exceeds $0.0044$). Following the Directionality Convention (Section~\ref{sec:method-group}), it is also worth noting the \emph{direction} of the nonzero gaps: five variants (V02, V03, V06, V08, V09) show an \emph{increase} in true-positive capture relative to \texttt{BASE} rather than a decrease, and only V04 shows a genuine decrease (a $2$-percentage-point true-positive-rate drop) -- so even the largest observed TPR gaps in this corpus are not predominantly evidence of qualified-candidate demotion.

This is the central empirical finding of the paper on this benchmark: on the same corpus, the same ranker, and the same nine bias variants, seven of the nine metric evaluations analyzed above -- score delta, flip rate, retention rate, impact ratio, recall, equal opportunity, and equalized odds -- report a uniformly clean audit for every treatment, while two more targeted signals still surface: MARC flags a single variant (V06) at the rank-stability boundary, and nDCG@$15$ flags three cells, including the neutral \texttt{BASE} configuration itself, on ranking-quality grounds unrelated to any demographic treatment. A practitioner relying only on score-delta, retention, or merit-aware rate-gap metrics -- the three families a four-fifths-rule-style audit most naturally reaches for -- would have certified this ranking behavior with a completely clean bill of health; only the rank-stability and ranking-quality families reveal that the picture is not perfectly uniform. This is direct empirical support for the paper's design decision (Section~\ref{sec:method-metrics}) to compute all three families rather than a single aggregate score, even though which family (if any) raises a flag is itself scale- and corpus-dependent, as Section~\ref{sec:results-discussion} discusses further.

\subsection{Discussion}
\label{sec:results-discussion}

Three points generalize beyond the specific numbers reported above. First, score stability is not rank stability. The cleanest empirical lesson of this illustrative run is the dissociation between Score Delta -- inside the $\pm0.02$ PASS band for every one of the nine variants (Section~\ref{sec:results-counterfactual}) -- and MARC, which averages $4.3$ to $5.3$ position-changes per variant. Even the variants with the smallest, most statistically unremarkable score shifts (V07 at $0.0025$, V09 at $0.0013$) still show more than four positions of average rank churn. Audit tooling that stopped at score parity would have certified this ranking behavior more confidently than the rank-level evidence supports: identity injection moved candidates several shortlist positions on average even where the underlying score shift was small, precisely because score perturbations translate into positional movement in a way a score-only audit cannot see, and at this corpus size that movement is large enough for one variant (V06) to cross the rank-stability INVESTIGATE boundary outright. For shortlist-mediated decisions, rank-stability metrics (MARC, Flip Rate, Retention Rate) should therefore be treated as first-class audit citizens rather than as secondary diagnostics to a score-level check. Second, the group-fairness and merit-aware rate-gap families happen to agree with each other on this corpus (Section~\ref{sec:results-group}), but nothing in their construction guarantees that: as \citet{hardt2016equality} establish for classifiers, aggregate outcome-rate parity and conditional (qualified-only) rate parity are formally distinct criteria that can diverge whenever a treatment's effect correlates with the underlying qualification signal, and any application of this or a similarly designed audit methodology should still compute both rather than infer one from the other. What this corpus does show diverging is coarser, threshold-crossing statistics (retention, impact ratio, equal opportunity, equalized odds -- each of which only asks whether a candidate is inside or outside a set, or matches or does not) versus within-list, position- and quality-sensitive statistics (MARC, nDCG): the former stay uniformly clean while the latter register the only two borderline findings in the entire report, precisely because set-membership statistics are blind to \emph{how much} movement occurs within the shortlist or to overall ranking-quality degradation. This is also the dual-invariant argument of Section~\ref{sec:method-stage1} manifesting empirically in a different guise: a system could satisfy every set-overlap and rate-gap invariant while still exhibiting position-level churn that only a rank- and quality-sensitive metric will catch. Third, any report produced by this methodology is most informative when read together with its $n$, CI width, and $p$-value columns rather than the PASS/INVESTIGATE/FAIL badge alone, particularly for boundary-adjacent findings such as V06's MARC result (Section~\ref{sec:results-artifact}), where the badge alone conveys none of the margin by which the finding was reached. Finally, the numerical-tolerance point of Section~\ref{sec:results-artifact} is a reminder that any fixed-threshold classification layer over continuous fairness statistics needs an explicit comparison tolerance built in, independent of how carefully the underlying metric formulas are derived.

\section{Conclusion and Future Discussion}
\label{sec:conclusion}

This paper presented a multi-agent LLM methodology that couples LLM-agent-driven, two-phase correspondence-audit generation -- an identity-neutral base-candidate phase followed by a bias-variant injection phase that structurally separates qualification signal from demographic signal -- with a nine-metric, three-family (counterfactual, group-fairness, merit-aware) quantitative fairness suite, each metric backed by a statistically appropriate significance test, a bootstrap confidence interval, and Benjamini--Hochberg false-discovery-rate correction, and reported through an automatically generated, threshold-classified HTML audit report with a composite weighted risk score. Illustrated on an example corpus spanning $5$ job orders, $100$ base candidates, and $10$ demographic-bias treatments, the methodology demonstrates two concrete findings that motivate its multi-metric design: (1)~seven of the nine metric evaluations -- including the family built to directly operationalize the legal four-fifths rule and the merit-aware rate-gap metrics -- report a uniformly clean audit ($0$ FAIL, $0$ INVESTIGATE) across every bias variant, while a rank-stability metric (MARC) flags one variant and a ranking-quality metric (nDCG@$K$) flags three cells, including the neutral baseline configuration itself, a divergence that would be invisible to any audit relying on a single aggregate fairness score; and (2)~a boundary-adjacent finding close to a PASS/INVESTIGATE threshold, together with a related floating-point numerical-tolerance hazard in the classification logic that can reclassify a boundary-exact effect size across a status boundary, are concrete reproducibility considerations for fixed-threshold audit classification generally.

Several limitations bound the scope of the illustrative results reported here and motivate future work. All reported numbers reflect a single experimental run of the example corpus with a locked configuration and a fixed random seed; no repeated trials are included, and borderline observations -- notably the four INVESTIGATE flags of Sections~\ref{sec:results-counterfactual} and~\ref{sec:results-merit}, including the boundary-adjacent MARC finding discussed in Section~\ref{sec:results-artifact} -- may be unstable under resampling of jobs, base candidates, or generation randomness. Repeated-trial validation is planned and is, in our view, required before results produced by this methodology are relied upon externally, including for legal purposes; the dual-threshold reporting design of Section~\ref{sec:method-dualthreshold} is deliberately not a substitute for that validation. The ground-truth job-relevance labels (\texttt{is\_true\_match}) underlying the merit-aware metrics are, in the example corpus, a designed alternating rule rather than recruiter-adjudicated labels, and in the underlying two-phase pipeline more generally, a self-reported \texttt{match} flag from the same Generation Agent that authors the resume; validating merit-aware findings against human-adjudicated relevance judgments on a larger corpus of real job orders is the most direct next step. The five-axis protected-characteristic taxonomy (Table~\ref{tab:axes}) and its automated LLM-based classification are themselves a modeling choice with unquantified classification error; an inter-rater reliability study comparing the automated classifier against human-labeled protected-axis annotations would bound how much of any observed fairness gap is attributable to axis-classification noise rather than genuine ranking-model behavior. The demographic-signal injection performed by the Generation Agent in Phase~B (Section~\ref{sec:method-stage1}) is itself an LLM generation step and could, in principle, introduce stereotyped or unrealistic framing when asked to "inject" a given axis; a human-review pass over a sample of generated bias variants, and/or cross-validation against hand-authored correspondence-audit resumes, would help establish how closely LLM-injected demographic signal matches the signal a human-constructed audit would use. Relatedly, LLM-generated resumes, however realistic, remain proxies for real candidate populations and may under-represent real-world confounds such as correlated attribute distributions, formatting diversity, and the many ways identity is signaled implicitly rather than declared; synthetic counterfactuals establish that a matching system \emph{can} respond to identity signals, but calibrating how those responses manifest on real traffic requires complementary production monitoring. At the statistical layer, any real deployment of this methodology will need substantially more accumulated candidate identities per job before its confidence intervals and significance tests are informative, particularly for the rank-and-set-based metrics that degenerate at small candidate-pool sizes relative to the shortlist size $K$ (every candidate is then trivially inside the top-$K$ under every treatment, and no rank-crossing is possible); systematically characterizing the minimum $n$ required for adequately powered bootstrap CIs and paired significance tests across the nine metrics, and building that guidance into the reporting layer itself (e.g., a minimum-$n$ warning before a report is generated), is a natural extension.

Two further ablations are planned that speak directly to the cross-family safeguards introduced in Sections~\ref{sec:lit-llm} and~\ref{sec:method-ranking}. First, swapping the generation and translation model to test whether generator-model choice itself introduces artifacts into the synthetic resumes, an effect the current design cannot distinguish from a genuine finding without such an ablation. Second, in deployments where the ranking step is itself LLM-mediated, varying evaluator capability and model family -- including a same-family generator/evaluator condition -- to test whether detected bias magnitude is a property of the ranking task or of the specific evaluator; a same-family evaluator reporting systematically lower bias on its own generations would quantify precisely the self-bias confound the cross-family design exists to avoid. Future work should also extend cross-validation of the merit-aware and group-fairness metrics against established fairness toolkits (Fairlearn, AIF360, Aequitas) reshaped into the paired-treatment structure this methodology assumes, to confirm that the custom implementations reported here converge with toolkit-computed analogues on overlapping definitions; broaden the protected-axis taxonomy and threshold calibration beyond the EU AI Act and the four-fifths rule to the other regional instruments introduced in Section~\ref{sec:method-axes} (Colorado's automated decision-making technology statute, New York City's Local Law 144, and California's automated-decision-making-technology regulations), validating that the same pipeline re-parameterizes cleanly rather than requiring a bespoke methodology per jurisdiction; extend the ranking-pipeline comparison to alternative embedding backbones (base versus fine-tuned) to determine whether fine-tuning that improves retrieval quality also changes the fairness profile reported here; generalize the illustrative deployment beyond a single job family to multiple job families, business units, and organizational brands, since nothing in the five-stage pipeline of Section~\ref{sec:methodology} is specific to any one of these; and add the explicit numerical tolerance to the classification logic recommended in Section~\ref{sec:results-artifact} to guard against floating-point threshold-boundary reclassification. Taken together, these steps would move the proposed methodology from a validated proof-of-concept -- as illustrated on the example corpus analyzed in this paper -- toward a continuously running fairness-audit component, open to external legal and regulatory review, deployable alongside any candidate--job matching pipeline it is used to evaluate.

\subsection*{Availability Statement}

The pipeline code, agent prompts, the example corpus, and the generated HTML audit report described in this paper are not released publicly alongside it; they remain internal artifacts of the deploying organization. They can, however, be made available for inspection, replication, or independent audit on a case-by-case basis upon direct request to the authors, subject to the deploying organization's review.

\bibliographystyle{unsrtnat}
\bibliography{references}

\end{document}